\PassOptionsToPackage{numbers,sort&compress}{natbib}

\documentclass{article}
\usepackage[preprint]{neurips_2025}

\usepackage[utf8]{inputenc}
\usepackage[T1]{fontenc}

\usepackage{graphicx}
\usepackage{subcaption}

\usepackage{amsfonts}
\usepackage{nicefrac}

\usepackage{booktabs}

\usepackage{microtype}

\usepackage{url}
\usepackage{hyperref}

\hypersetup{
    colorlinks,
    linkcolor={red!50!black},
    citecolor={blue!50!black},
    urlcolor={blue!80!black}
}

\usepackage{microtype}
\usepackage{graphicx}
\usepackage{subcaption}
\usepackage{booktabs}
\usepackage[dvipsnames]{xcolor}
\usepackage{caption}
\usepackage{wrapfig}
\usepackage{graphics}
\usepackage{enumitem}
\usepackage{amsmath}
\usepackage{amssymb}
\usepackage{mathtools}
\usepackage{amsthm}
\usepackage[skip=10pt plus1pt, indent=0pt]{parskip}
\let\svthefootnote\thefootnote
\newcommand\freefootnote[1]{%
  \let\thefootnote\relax%
  \footnotetext{#1}%
  \let\thefootnote\svthefootnote%
}
\usepackage{multirow}
\usepackage{algorithm}
\usepackage{algpseudocode}
\usepackage{duckuments}
\usepackage{pifont}
\usepackage{tcolorbox}
\tcbuselibrary{skins, breakable, theorems}
\usepackage[titletoc]{appendix}
\usepackage{listings}
\newtcolorbox{promptbox}[1][]{
  colback=gray!5,
  colframe=gray!50,
  coltitle=black,
  colbacktitle=yellow!20,
  fonttitle=\bfseries,
  title=Prompt,
  enhanced,
  sharp corners=south,
  boxrule=0.8pt,
  left=6pt,
  right=6pt,
  top=6pt,
  bottom=6pt,
  width=\linewidth,
  fontupper=\ttfamily\small,
  before skip=6pt,
  after skip=6pt,
  #1
}

\newtcolorbox{llmbox}[1][]{
  colback=gray!5,
  colframe=gray!50,
  coltitle=black,
  colbacktitle=blue!15,     
  fonttitle=\bfseries,
  title=LLM Output,
  enhanced,
  sharp corners=south,
  boxrule=0.8pt,
  left=6pt,
  right=6pt,
  top=6pt,
  bottom=6pt,
  width=\linewidth,
  fontupper=\ttfamily\small,
  before skip=6pt,
  after skip=6pt,
  #1
}
\newtcolorbox{llmbox2}[1][]{
  colback=gray!5,
  colframe=gray!50,
  coltitle=black,
  colbacktitle=green!15,
  fonttitle=\bfseries,
  title=LLM Output (Alt),
  enhanced,
  sharp corners=south,
  boxrule=0.8pt,
  left=6pt,
  right=6pt,
  top=6pt,
  bottom=6pt,
  width=\linewidth,
  fontupper=\ttfamily\small,
  before skip=6pt,
  after skip=6pt,
  #1
}
\newtcolorbox{outputbopx}[1][]{
  colback=gray!5,
  colframe=gray!50,
  coltitle=black,
  colbacktitle=red!15,
  fonttitle=\bfseries,
  title=LLM Output (Alt),
  enhanced,
  sharp corners=south,
  boxrule=0.8pt,
  left=6pt,
  right=6pt,
  top=6pt,
  bottom=6pt,
  width=\linewidth,
  fontupper=\ttfamily\small,
  before skip=6pt,
  after skip=6pt,
  #1
}

\makeatletter
\let\blx@rerun@biber\relax
\makeatother

\renewcommand{\vec}[1]{\ensuremath{\boldsymbol{#1}}}

\makeatletter
\let\blx@rerun@biber\relax
\makeatother

\newtcolorbox{definitionbox}[1][]{
  enhanced,
  colback=gray!10,
  colframe=gray!80,
  coltitle=black,
  fonttitle=\bfseries,
  title=Definition,
  #1
}

\usepackage{lineno}

\definecolor{darkblue}{rgb}{0, 0, 0.5}
\hypersetup{colorlinks=true, citecolor=darkblue, linkcolor=darkblue, urlcolor=darkblue}

\title{
Decomposing Elements of Problem Solving: \\
What "Math" Does RL Teach?
}
\author{%
  Tian Qin\textsuperscript{1}\thanks{Equal contributions. Correspondence to \texttt{tqin@g.harvard.edu}, \texttt{corefranciscopark@g.harvard.edu}} \\
  \And
  Core Francisco Park\textsuperscript{1}\footnotemark[1] \\
  \AND
  Mujin Kwun\textsuperscript{1,2}\\
  \And
  Aaron Walsman\textsuperscript{1,2}\\
  \And
  Eran Malach\textsuperscript{1,2}\\
  \And
  Nikhil Anand\textsuperscript{1,2}\\
  \AND
  Hidenori Tanaka\textsuperscript{1}\thanks{Co-senior authors.} \\
  \And
  David Alvarez-Melis\textsuperscript{1,2}\footnotemark[2] 
}

\begin{document}

\maketitle
\begin{center}
    \vspace{-20px}
    \textsuperscript{1}Harvard University \quad \textsuperscript{2}Kempner Institute
    \vspace{20px}
\end{center}

\begin{abstract}
    Mathematical reasoning tasks have become prominent benchmarks for assessing the reasoning capabilities of LLMs, especially with reinforcement learning (RL) methods such as GRPO showing significant performance gains. However, accuracy metrics alone do not support fine-grained assessment of capabilities and fail to reveal which problem-solving skills have been internalized. To better understand these capabilities, 
    we propose to decompose problem solving into fundamental 
    capabilities: \textit{Plan} (mapping questions to sequences of steps), \textit{Execute} (correctly performing solution steps), and \textit{Verify} (identifying the correctness of a solution). Empirically, we find that GRPO mainly enhances the \textit{execution} skill---improving execution robustness on problems the model already knows how to solve---a phenomenon we call \textit{temperature distillation}. 
    More importantly, we show that RL-trained models struggle with fundamentally new problems, hitting a `coverage wall' due to insufficient \textit{planning} skills. 
    To explore RL's impact more deeply, we construct a minimal, synthetic solution-tree navigation task as an analogy for mathematical problem-solving. 
    This controlled setup replicates our empirical findings, confirming RL primarily boosts execution robustness.
    Importantly, in this setting, we identify conditions under which RL can potentially overcome the coverage wall through improved exploration and generalization to new solution paths. 
    Our findings provide insights into the role of RL in enhancing LLM reasoning, expose key limitations, and suggest a path toward overcoming these barriers.
    Code is available at \url{https://github.com/cfpark00/RL-Wall}.
\end{abstract}

\section{Introduction}

\begin{figure}
    \centering
    \includegraphics[width=0.95\linewidth]{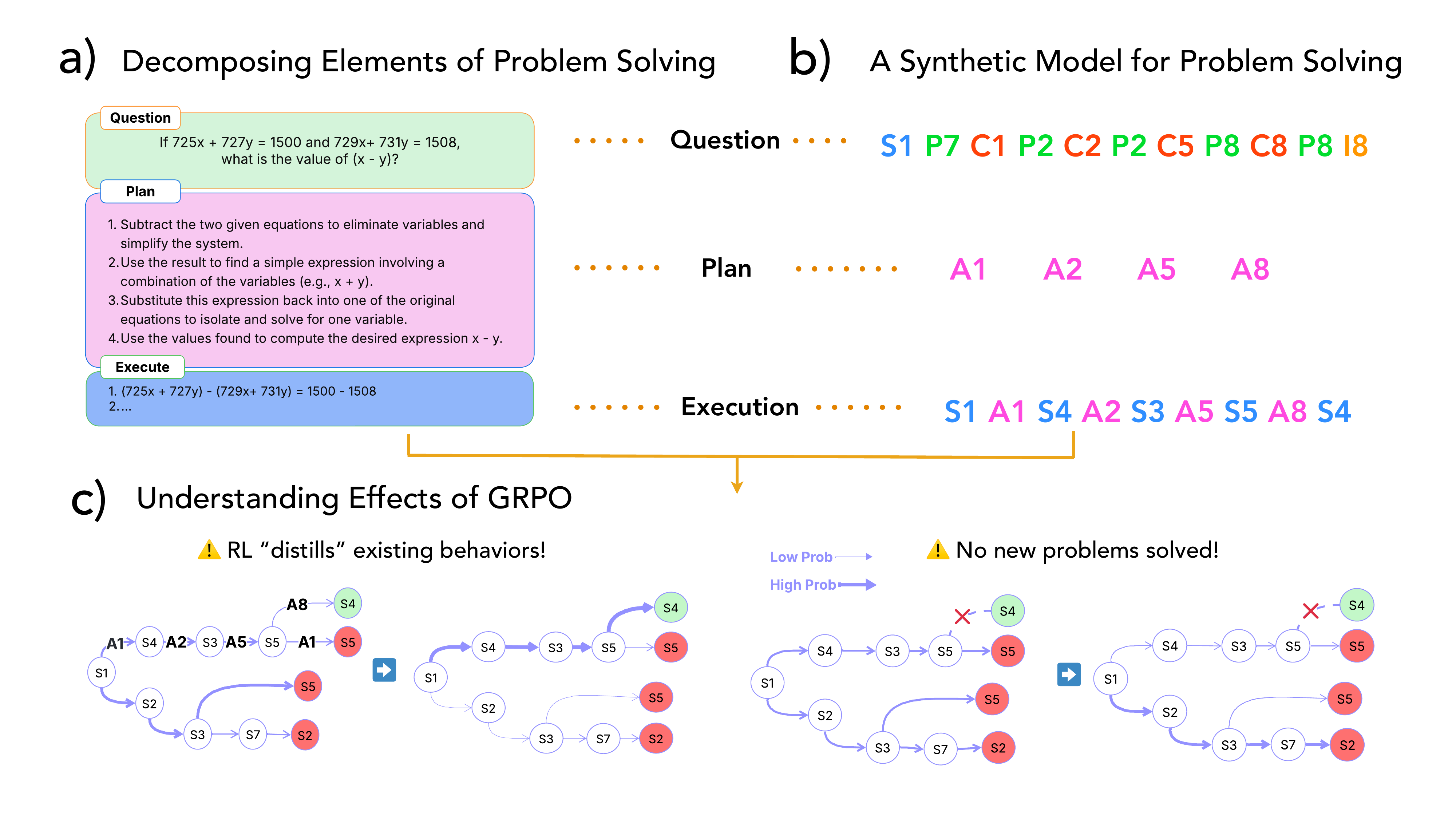}
    \caption{
    \textbf{Decomposing problem solving and modeling the effects of RL in a toy setting.} 
    \textit{(a):} We decompose math problem solving into fundamental capabilities. To solve a problem model must first \textit{plan} (map the question to a sequence of actions) and then execute (navigate the tree) without errors.
    \textit{(b):} We design a minimal toy environment that explicitly captures these core reasoning elements in a controlled setting.
    \textit{(c):}  GRPO primarily improves precision on problems the model already partially solves, without expanding overall \textit{coverage}. Our synthetic setup reproduces this behavior—and crucially, shows that under the right conditions, RL \textit{can} break the coverage wall.
    }\label{fig:main}
\end{figure}

Recent months have seen rapid gains in LLM mathematical reasoning, driven in large part by reinforcement learning (RL) methods like the simple yet effective GRPO \cite{Shao2024-me}. This progress has largely been benchmarked using a single evaluation metric: Pass@1 accuracy---the fraction of problems for which the model produces a correct solution on its first attempt. This mirrors a long-standing tendency in human cognition research to rely on simple, aggregate metrics---such as standardized test scores \cite{Sackett2008-bf, Au2007-ea} or IQ measures\cite{Gottfredson1997-eb}---to evaluate complex capabilities. These metrics persist in part because they are easy to interpret and compare across individuals \cite{Gigerenzer2011-cp, Kahneman2011-ni}. While convenient, relying on a single accuracy score like Pass@1 to evaluate LLM reasoning has two important limitations. 
First, it is \textit{monolithic}: too coarse-grained to reveal the nature of model failures or the specific reasoning skills being improved, such as whether RL enhances planning, computation, error-checking, or merely reinforces existing knowledge. Second, it evaluates greedy decoding, whereas real-world deployments---especially those requiring complex reasoning---often rely on stochastic sampling (e.g., temperature decoding, best-of-N selection) to increase diversity or robustness. To truly understand the impact of RL on reasoning, we need evaluation methods that go beyond aggregate accuracy and capture how models navigate the problem-solving process.

To address these limitations, we adopt a finer-grained evaluation framework aimed at understanding two key questions: (i) are the improvements from GRPO uniform across problems?, and (ii) what specific aspects of reasoning does GRPO improve?
By breaking down accuracy per problem, we find that GRPO is most effective on problems the model already knows how to solve  (Section \ref{sec:fine_grained}). On these problems, its effect is to make the correct solution more likely regardless of sampling temperature---a phenomenon we call \textit{temperature distillation}. This enhances execution robustness under sampling but does little to help with fundamentally novel problems, exposing a \textit{coverage wall}, defined as the limit of Pass@$K$ as $K \rightarrow \infty$ (Fig.~\ref{fig:main}\textit{c}). To study (ii) we introduce a more fine-grained framework for analyzing reasoning, viewing problem solving as a search process in a solution tree (Section \ref{sec:elements}). This perspective decomposes reasoning into three core skills (Fig.~\ref{fig:main}\textit{a}): Plan (mapping the question to a sequence of actions indicating where to go in the tree, Execute (carrying out those steps correctly by performing the navigation), and Verify (detecting and correcting errors). Using this lens, we find that even small models often plan well but frequently fail in execution, making logical and factual errors. GRPO, we show, primarily strengthens execution.

To better isolate and understand RL’s effects, we construct a minimal, synthetic setting that abstracts mathematical problem solving. (Fig.~\ref{fig:main}\textit{b}, Sec.~\ref{sec:syn_math}). Here, the model needs to map the question to a sequence of actions (i.e., plan) and execute the sequence of actions to provide the correct answer. This controlled environment closely mirrors our proposed framework, while allowing control over the underlying data generation process. We find that this synthetic setting faithfully reproduces our empirical observations. Crucially, the setup also reveals properties of the pretraining and RL data which allows RL to break through the coverage wall---namely, when the model is able to explore new actions and apply them to structurally similar problems. 

Our study provides insights into RL's role in enhancing LLM reasoning. Our contributions are:
\vspace{-0.2cm}
\begin{itemize}[itemsep=2pt,labelindent=2pt,topsep=0pt,parsep=0pt,partopsep=0pt, align=left, leftmargin=*]
\item \textbf{Reasoning Decomposition:} We propose a framework that decomposes math problem solving into three capabilities: planning, execution, and verification. 
\item \textbf{Empirical Analysis of RL:} In the proposed decomposition, we show that GRPO primarily improves execution on known problems, leading to a temperature distillation effect. However, GRPO fails to help model solve previously unsolved problems, revealing a coverage wall.
\item \textbf{Synthetic Validation:} We construct a minimal synthetic task that closely reflects the fundamental elements of problem solving and replicates empirical findings. We reveal some data properties under which RL can overcome the coverage wall.
\end{itemize}

\section{Related Work}
\subsection{Math Reasoning Benchmarks}
Mathematics has long been a core domain for evaluating AI reasoning, dating back to early work on algebra word problems and logic inference \citep{Bobrow1964-mb, Koncel-Kedziorski2016-zm, Wang2017-xj, Amini2019-uc}. Recently, math word problems involving basic arithmetic and multi-step reasoning such as GSM8K \citep{Cobbe2021-ag} and MATH \citep{Hendrycks2021-nv} has become standard benchmarks for elementary and high school–level tasks \citep{Lu2023-uh}. As model performance has saturated on many of these benchmarks, recent efforts have pushed toward more challenging datasets include olympiad- and competition-level problems \citep{He2024-xl, Tsoukalas2024-ox, Gao2024-hh}. Outside math word problems, the reasoning capabilities have been extended to theorem proving benchmarks \citep{Zheng2021-hb, Welleck2021-sf} as well as multi-modal problems \citep{Lu2022-if, Chen2021-xi, Yue2023-xb}. While benchmark complexity has increased, relatively little work has focused on understanding what these benchmarks actually measure about LLM reasoning—and which skills drive model performance improvements.

\subsection{Reinforcement Learning for LLM Post Training}
Post-training methods for improving LLM reasoning generally fall into three categories: supervised fine-tuning \citep{Ouyang2022-lv, Zelikman2022-yo}, self-improvement \citep{Wang2023-wv, Gulcehre2023-hq}, and reinforcement learning (RL). Within RL, PPO and its variants \citep{Schulman2017-tw, Kazemnejad2024-gw, Yue2025-gg} rely on training a critic model to produce reward signals. In contrast, Reinforcement Learning with Verifiable Rewards (RLVR) has gained prominence in domains like mathematics and programming, where rule-based verification enables automatic reward generation \citep{Lambert2024-go, Shao2024-me}. The most common reward design uses simple answer matching \citep{Kimi-Team2025-hf, DeepSeek-AI2025-vs, Zeng2025-xn, Wen2025-ma}—a binary signal that circumvents the need for complex process-based rewards \citep{Lightman2023-cg}.
Recent large-scale applications of RLVR, such as OpenAI’s o1 \citep{OpenAI2024-cj} and DeepSeek-R1 \citep{DeepSeek-AI2025-vs}, have demonstrated its potential to scale reasoning capabilities in LLMs. Notably, DeepSeek-R1 introduced the influential zero RL paradigm, where reinforcement learning is applied directly to a base model without any intermediate supervised tuning. This approach has shown that RL alone can induce long-chain chain-of-thought (CoT) reasoning and self-reflective behavior \citep{DeepSeek-AI2025-vs, Zeng2025-xn}, sparking a wave of open-source replications and algorithmic advances \citep{Liu2025-gj, Luo2025-ql, Shen2025-fv, Zhang2025-qp, Zhao2025-qc, Wang2025-zg}.

While these efforts aim to improve final-task accuracy, our work takes a complementary perspective: examining which elementary reasoning skills are improved through RL to identify where fundamental limitations remain. Concurrent work has also observed that GRPO does not improve coverage \citep{Yue2025-se, Dang2025-zz}. While their results focuses on confirming this phenomenon across training regimes, datasets, and model scales, our work emphasizes on diagnosing the underlying cause. Additionally, we construct a toy model of problem solving, enabling controlled investigation RL's effect.


\section{Breaking Down the Accuracy Metric}
\label{sec:fine_grained}

In this section, we first argue that a single accuracy score is insufficient to comprehensively evaluate model performance (Section \ref{sec:beyond_accuracy}). We then show that moving to a more fine-grained metric reveals that the predominant effect of RL can be understood as \textit{temperature distillation} (an increase in precision, Section \ref{sec:temp_distill}) without improving coverage (Section \ref{sec:coverage}). 

\subsection{Going Beyond Accuracy}
\label{sec:beyond_accuracy}
\begin{figure}[t]
    \centering
    \includegraphics[width=0.9\linewidth]{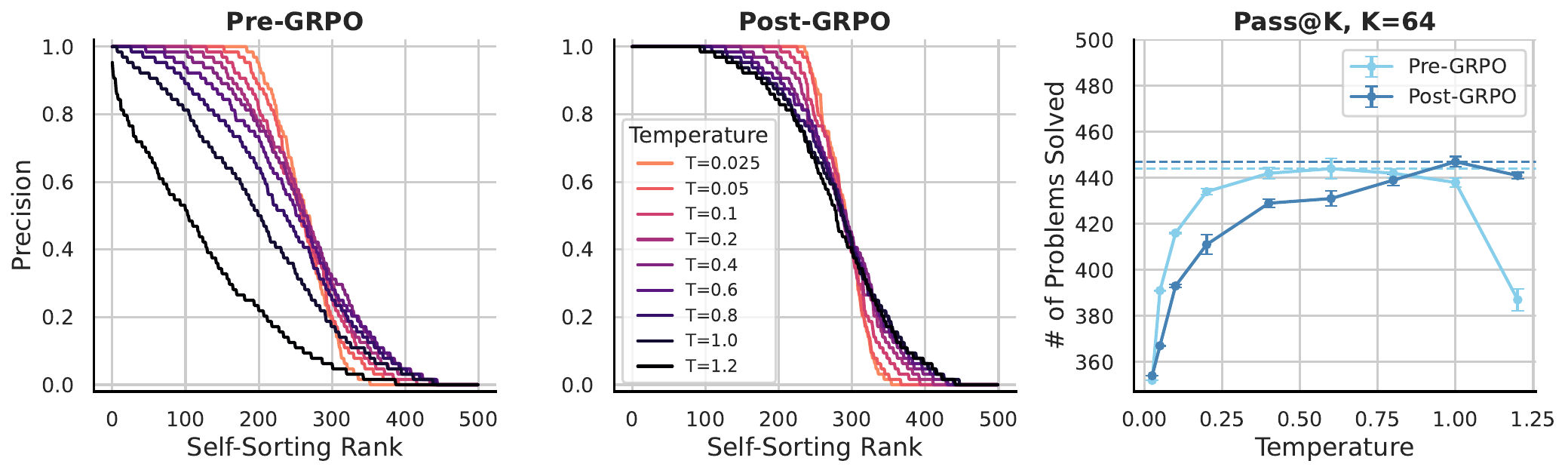}
    \caption{
    \textbf{GRPO improves precision (average@$K$) through temperature distillation but does not improve coverage (measured as pass@$K$).}
    \textit{Left, Middle:} Precision (probability of correctly solving the problem) for MATH-500 questions before and after performing GRPO on Qwen2.5-1.5B-Instruct. GRPO ``tightens" the model's behaviors, which we study as the ``temperature distillation effect".   
    \textit{Right:} Pass@64 at various temperature settings (\textit{solid lines}). We define coverage as the best Pass@64 rate over temperatures (\textit{dashed lines}). GRPO does not improve coverage. 
    }
    \label{fig:qwen_temp}
\end{figure}

Recent progress in LLM reasoning---especially using RL---is often reported using a single metric: Pass@1 at a fixed sampling temperature \citep{Liu2025-gj, DeepSeek-AI2025-vs, Zeng2025-xn}. With models surpassing 70\% accuracy on hard math benchmarks, it’s tempting to conclude they are capable reasoners. However, this metric can be misleading. First, real-world deployments rarely operate at a fixed temperature; robust reasoning should generalize across sampling regimes. Second, many use cases rely on test-time scaling strategies like Best-of-N \citep{Stiennon2020-nm} or Majority Voting \citep{Wang2022-or}, which Pass@K (especially with $K=1$ at a fixed temperature) is insensitive to. As a result, we have limited insights into what capabilities RL actually enhances---a gap which our experimental setup seeks to address. 

\paragraph{Model Training.} For this, we train Qwen2.5-Instruct models (0.5B, 1.5B, 7B) \citep{Team2024-qy} on MATH \citep{Hendrycks2021-nv} using GRPO with the VERL \citep{Sheng2024-rs} and evaluate on MATH-500 \citep{Hendrycks2021-nv}. We closely follow the setup in \citet{Zeng2025-xn} that reproduces the DeepSeek-R1 pipeline \citep{DeepSeek-AI2025-vs}. We then examine what those gains reveal about actual problem-solving ability. See additional experiment detail in Appendix \ref{appx:experimental_details}.

\subsection{GRPO Distills Temperature} 
\label{sec:temp_distill} 
To assess model performance beyond Pass@1, we sample 64 completions per test problem across a range of temperatures ($0 \leq T \leq 1.2$), both before and after GRPO. For each problem, we compute its \textit{precision}: the fraction of correct generations. To visualize trends, we rank problems by their own precision scores---what we call \textit{self-difficulty sorting} (or \textit{self-sorting})---allowing us to analyze how temperature affects performance across problems of varying difficulty.
Fig.~\ref{fig:qwen_temp} (\textit{left, middle}) shows the precision curves for Qwen2.5-1.5B-Instruct.\footnote{We focus on Qwen2.5-1.5B-Instruct in the main paper. See Appendix \ref{appdx:additional_eval} for valuations across different model sizes, training data and test benchmarks.} Before GRPO, temperature heavily affects performance: higher temperatures reduce precision on easy problems but slightly improve success on harder ones, reflecting more exploratory behavior. After GRPO, this temperature sensitivity largely disappears—precision curves flatten across all temperatures. 
We refer to this robustness to sampling temperature as the \textit{temperature distillation effect}. This phenomenon is only partially reflected using Pass@$K$, which captures model behavior at a fixed temperature and overlooks full distribution change across the problem difficulty spectrum.

\subsection{GRPO Does Not Unlock Previously Unsolvable Problems}
\label{sec:coverage}
A central question we examine is whether reinforcement learning enables the model to solve problems it previously could not. We address this using two complementary approaches: total coverage and fine-grained matched problem analysis.

\vspace{-0.1cm}
\paragraph{Coverage.} We define $\text{Coverage}(K, T^*)$ as the number of problems for which at least one of the $K$ samples generated at optimal temperature $T^*$ is correct. 
In practice, the number of solvable problems grows roughly log-linearly with $K$ \citep{Brown2024-ul}, 
so $K$ is typically chosen as large as computational constraints allow based on task difficulty. Here we use $K = 64$ for the MATH-500 (see Appendix~\ref{appdx:coverage_k} for justification). To isolate coverage from temperature sensitivity---a key confounder due to the temperature distillation effect---coverage requires the optimal sampling temperature for each model. To account sampling stochasticity, we run best-of-64 sampling across 5 random seeds for each problem, both before and after GRPO. Fig.~\ref{fig:qwen_temp} (\textit{right}) shows the mean and standard deviation (solid lines), as well as the number of problems solved (coverage) at each model's optimal sampling temperature (dashed lines). The key observation is that GRPO does not significantly increase coverage. However, coverage alone offers an incomplete view: it doesn't reveal \textit{which} problems are being solved. Two models with similar coverage could succeed on entirely different subsets, implying a trade-off in problem-solving behavior. We explore this further by analyzing per-problem solution rates.\looseness=-1

\begin{figure}[t]
    \centering
    \includegraphics[width=0.42\linewidth]{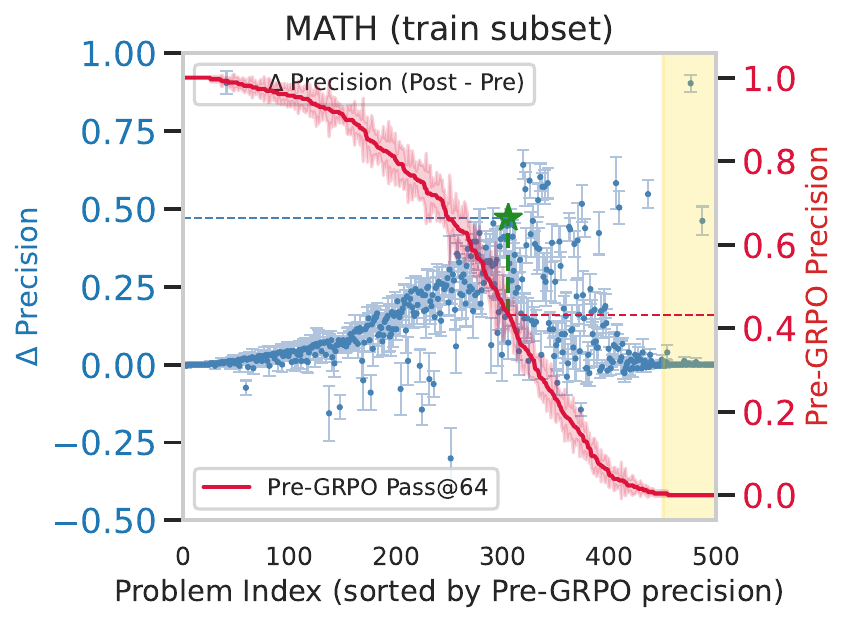}
    \hspace{10px}
    \includegraphics[width=0.42\linewidth]{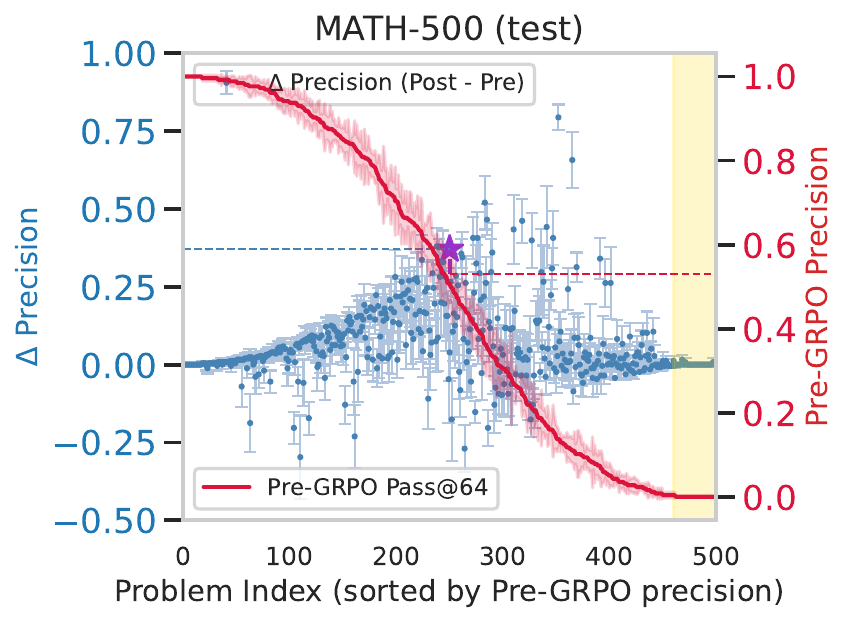}
    \caption{
    \textbf{Per-problem precision improvement on train subset (\textit{left}) and test set (\textit{right}).} We sort problems by pre-GRPO model's precision, and examine the improvement from running GRPO. We report mean precision and standard deviations from 5 random seeds. 
    GRPO improves precision on questions of medium-level difficulty while encountering a train-test generalization gap (\textit{green, purple star}).
    GRPO also encounters a coverage wall (\textit{yellow}) on hard problems. 
    }\label{fig:qwen_macth_problem}
\end{figure}

\vspace{-0.1cm}
\paragraph{Per-problem Improvement and Generalization Gap.} 
To study fine-grained changes in problem solvability from GRPO, we conduct a \textit{matched-problem} analysis: problems are sorted by their pre-GRPO precision, and we directly compare pre- and post-GRPO precision for each.  Fig.~\ref{fig:qwen_macth_problem} (\textit{left}) shows results on a subset of the training set; Fig.~\ref{fig:qwen_macth_problem} (\textit{right}) shows test set results. for problems the model already solves reliably (high pre-GRPO precision), GRPO yields minimial gains due to performance saturation. Improvements are concentrated in problems with medium initial precision. Interestingly, we observe a clear train–test generalization gap in this medium-precision regime. On the training set (Fig.\ref{fig:qwen_macth_problem}, \textit{left}), problems with 40\% pre-GRPO precision (\textit{green star}) see the largest improvements, with an impressive 45\% absolute gain. In contrast, on the test set (Fig.\ref{fig:qwen_macth_problem}, \textit{right}), the peak improvement shifts to problems with higher ($\sim60\%$) pre-GRPO precision (\textit{purple}), where the model could already succeed using majority voting. Moreover, the maximum gain is smaller---around 35\%. Beyond this medium-precision zone, gains drop sharply. On the training subset, GRPO unlocks only two new problems. Crucially, these gains do not generalize: on the test set, GRPO fails to solve any new problems, revealing what we term a \textit{coverage wall}.

\section{Elements of Problem Solving}
\label{sec:elements}
So far, we’ve seen that RL primarily results in temperature distillation and fails to overcome the coverage wall. To understand why, we decompose math problem solving into three core components (Section~\ref{sec:plan_execute}) and use this framework to analyze how RL shapes model behavior (Sections~\ref{sec:direction_execution} and~\ref{sec:execution_rubric}).

\subsection{Plan, Execute, and Verify}
\label{sec:plan_execute}
We conceptualize math problem solving as traversing through a solution tree, where each node represents a possible intermediate step. Within this framework, we identify three essential capabilities:
\vspace{-0.1cm}
\begin{itemize}[itemsep=1pt,labelindent=2pt,topsep=0pt,parsep=0pt,partopsep=0pt, align=left, leftmargin=*]
    \item Plan: Mapping a problem to a sequence of solution steps indicating where to go in the tree. Planning is non-trivial most problems don’t explicitly indicate the necessary steps to solve them.
    \item Execute: Correctly carrying out these steps, which requires mathematical knowledge, contextual reasoning, and logical consistency.
    \item Verify: Detecting and correcting errors in the solution path. While large models can sometimes backtrack or reflect, such behaviors are limited in smaller models. Verification is an active area of recent research \citep{Wang2024-up, Havrilla2024-bl, Gandhi2025-ac}.
\end{itemize}
Here we focus on the Plan and Execute components. Our goal is to identify where models succeed or fail in these steps and, most importantly, whether RL enhances these abilities.

\subsection{Easy To Plan But Hard To Execute}
\label{sec:direction_execution}

\begin{figure}[t]
    \centering
    \includegraphics[height=0.42\linewidth]{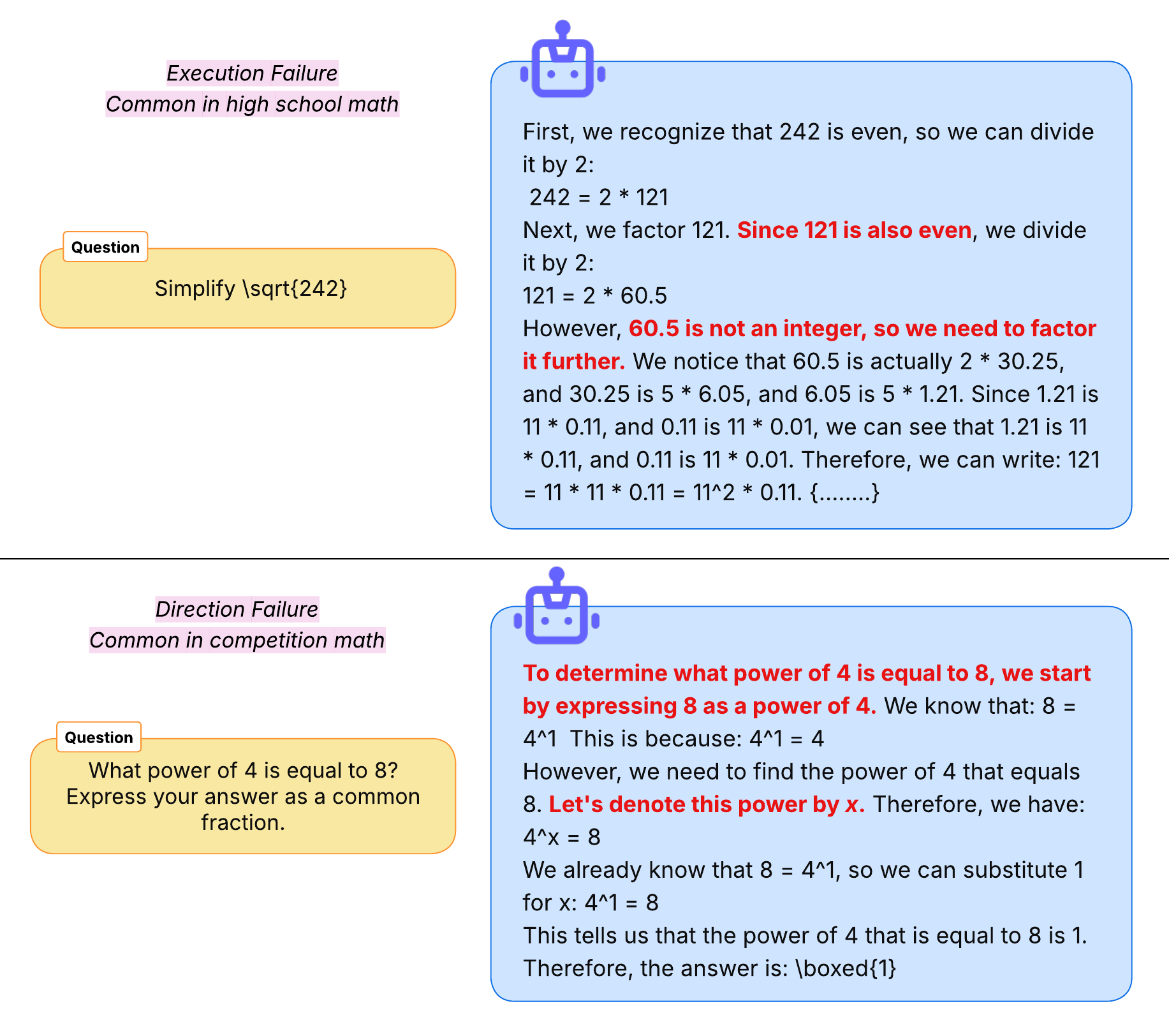}
    \hspace{10px}
    \includegraphics[height=0.42\linewidth]{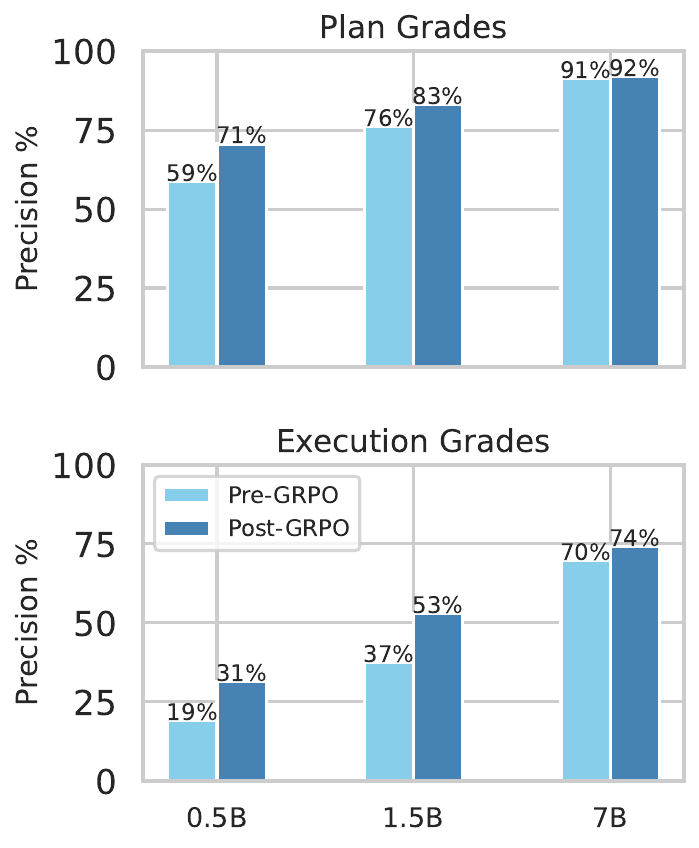}
    \caption{
    \textbf{Two elements to ace a math exam: Plan then Execute.}
    \textit{Left:} Examples of planning and execution failures from Qwen2.5-1.5B-Instruct on MATH-500 questions. 
    \textit{Right:} For MATH-500, even a small model knows how to approach (i.e., \textit{plan}) most problems. However, the ability to correctly \textit{executing} them depends heavily on model sizes. GRPO has a larger impact on a model's execution ability (i.e., improving \textit{execution robustness}). 
    }
    \label{fig:human_annot}
\end{figure}

\paragraph{Methodology.} To analyze model behavior, we first check whether each generated solution contains a valid plan---that is, whether the trace contains the correct approach. 
For challenging tasks such as Olympiad or research math, there are likely different approaches to the same problem. However, for high school level questions such as MATH-500, the solution approaches are mostly unique and involves 3 to 4 steps on average.
We automate the annotation process using GPT-4.1-mini as follows.\footnote{\texttt{gpt-4.1-mini-2025-04-14}. See Appendix~\ref{appdx:human_annot_example} for prompt details.} We begin by summarizing the ground-truth solution, listing the key steps required to solve the problem, without working out the math operations. Fig.~\ref{fig:main}(\textit{a}) illustrates step plan for a question from MATH-500. GPT then checks whether the model's solution trace includes all essential steps. If the plan is valid, we next assess execution: does the model carry out the steps correctly? Again, we treat this as a classification task and ask GPT to check for any mathematical or logical errors leading to an incorrect final answer. In Appendix~\ref{appdx:human_annot_example}, we list 20 randomly selected annotated examples. Manual inspection confirms that GPT's annotations align closely with human judgment. 
In Fig.~\ref{fig:human_annot} (\textit{right}), we list two examples of direction and execution failure made by 1.5B model on MATH-500. \footnote{The correct plan for the plan failure question starts with ``express both sides of the equation with the same base and then equate the exponents." }

\paragraph{Results.} Using questions from MATH-500, we evaluate 64 generations per problem sampled at $T{=}1$. We compute the \textbf{plan grade}: the fraction of generations that include all essential steps from the reference summary. Among question with correct plan grade, the \textbf{execution grade} then measures the fraction for which the steps are carried out correctly. As shown in Fig.~\ref{fig:human_annot} (\textit{top right}), even the smallest model (0.5B) identifies valid solution plans for most problems---suggesting that to solve high school math, large-scale pretraining enables reliable planning, even in smaller models. We suspect that plan will be much harder for advanced competition level math. In contrast, execution shows much steeper scaling (Fig.~\ref{fig:human_annot}, \textit{bottom right}): larger models are far better at executing plans correctly. This points to execution, not planning, as the primary bottleneck in high math problem solving. Notably, GRPO improves execution robustness across all model sizes, with a smaller impact on planning. 

\begin{figure}[t]
    \centering
    \includegraphics[height=0.32\linewidth]{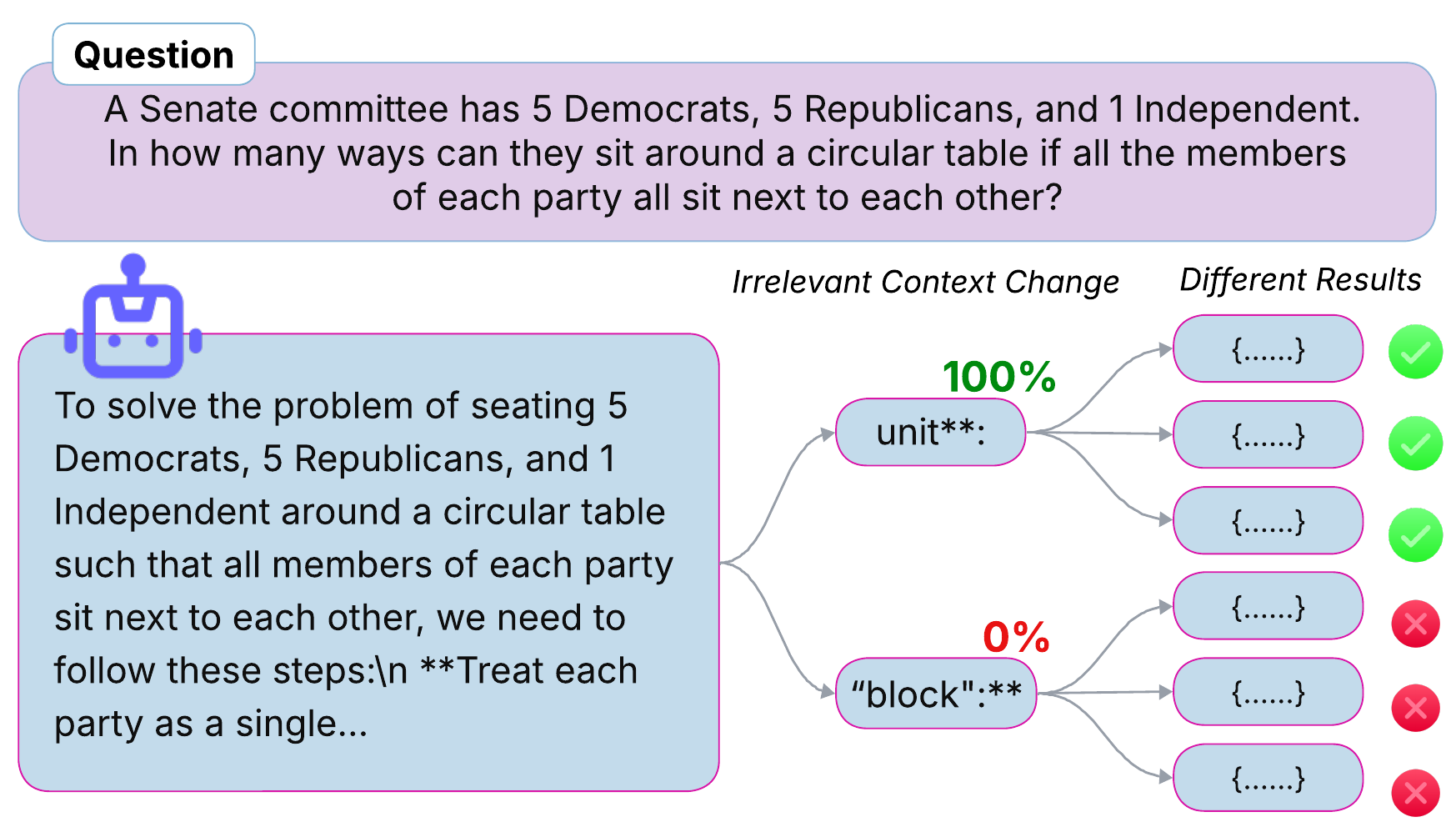}
    \includegraphics[height=0.3\linewidth]{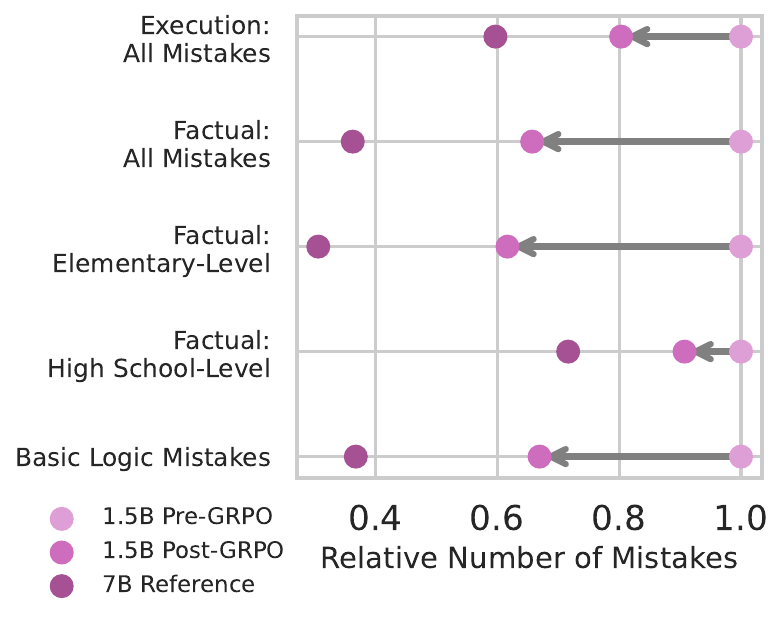}
    \caption{
    \textbf{Model cannot reliably execute math and logic reasoning due to spurious correlations.}
    \textit{Left:} One example of solution tree from Qwen2.5-1.5B-Instruct on a MATH-500 question. In this question, the model's likelihood of outputting the correct solution rely heavily on a semantically irrelevant context (saying ``unit" versus ``block").  
    \textit{Right:} To examinbe whether GRPO fixes spurious correlations, we annotate all the generations with execution failures (Figure \ref{fig:human_annot}), and examine the relative number of mistakes before and after GRPO: GRPO helps model fixes basic math and logic errors but does not improve advanced (high school) math. 
    }
    \label{fig:exec_mistakes}
\end{figure}

\subsection{Reasoning with Spurious Correlations}
\label{sec:execution_rubric}

\paragraph{Spurious Correlations.}
Why do models often fail to execute solution steps correctly? Understanding these failures would shed light on the limits of their reasoning capabilities. To this end, we implement a visualization tool (See Appendix \ref{appdx:solution_tree}) that convert model's solution roll out into a solution tree, branching whenever generations diverge. Using this visualization tool, we inspect model's solution tree across 30 different MATH-500 questions. A consistent pattern emerges: models apply mathematical and logical steps \textit{probabilistically}, often influenced by spurious correlations such as formatting or phrasing in the chain-of-thought. For instance, Fig.~\ref{fig:exec_mistakes} (\textit{left}) shows Qwen2.5-1.5B’s generation tree on a problem where solutions diverge based on whether the word ``unit" or ``block" is used. Although the wording choice is semantically irrelevant, it significantly affects whether the model produces a correct solution. This suggests that execution accuracy is driven more by superficial patterns in the training data than by robust reasoning. Despite strong benchmark scores (1.5B achieves $\geq50\%$ on high shcool MATH-500, $\geq70\%$ on grade-school GSM8K)---these models still make basic factual and logical errors---many of which would be obvious to elementary school students (see additional examples in Appendix~\ref{appdx:solution_tree}). We hypothesize that temperature distillation mitigates these issues by encouraging solution traces that are less sensitive to such spurious cues, improving consistency across generations.

\paragraph{Partial Repair via GRPO.} 
To test this hypothesis, we re-visit solution traces previously identified as containing execution errors (Section~\ref{sec:direction_execution}) and label them with:
\vspace{-0.1cm}
\begin{enumerate}[itemsep=0pt,labelindent=2pt,topsep=0pt,parsep=0pt,partopsep=0pt, align=left, leftmargin=*]
    \item whether they contain a basic factual math error (e.g., claiming that a triangle has six sides)
    \item the difficulty level of the mistaken fact (elementary or high-school level)
    \item whether there is a basic logical error (e.g., a contradiction or a \textit{non-sequitur})
\end{enumerate}
\vspace{-0.1cm}
We use GPT-4.1-mini to generate these annotations pre- and post-GRPO. Appendix~\ref{appdx:rubric_details} includes annotation prompts and sample outputs. Fig.~\ref{fig:exec_mistakes} (\textit{right}) shows the relative reduction in error rates, normalized to the number of pre-GRPO mistakes (\textit{light purple}). GRPO significantly reduces both elementary-level math errors and basic logic mistakes (\textit{purple}), improving robustness to spurious cues. For reference, we include a larger 7B model's performance (\textit{dark purple}). However, GRPO does not meaningfully reduce high-school-level factual errors, suggesting it enhances stability rather than teaching new mathematical knowledge. 

Taken together, these results reinforce and provide an explanation for our earlier findings: GRPO improves execution be reducing low-level errors and increasing internal consistency---but it does not enhance planning or expand mathematical understanding.  

\section{A Toy Model for Problem Solving}
\label{sec:syn_math}

\begin{figure}[!h]
    \centering
    \includegraphics[width=0.85\linewidth]{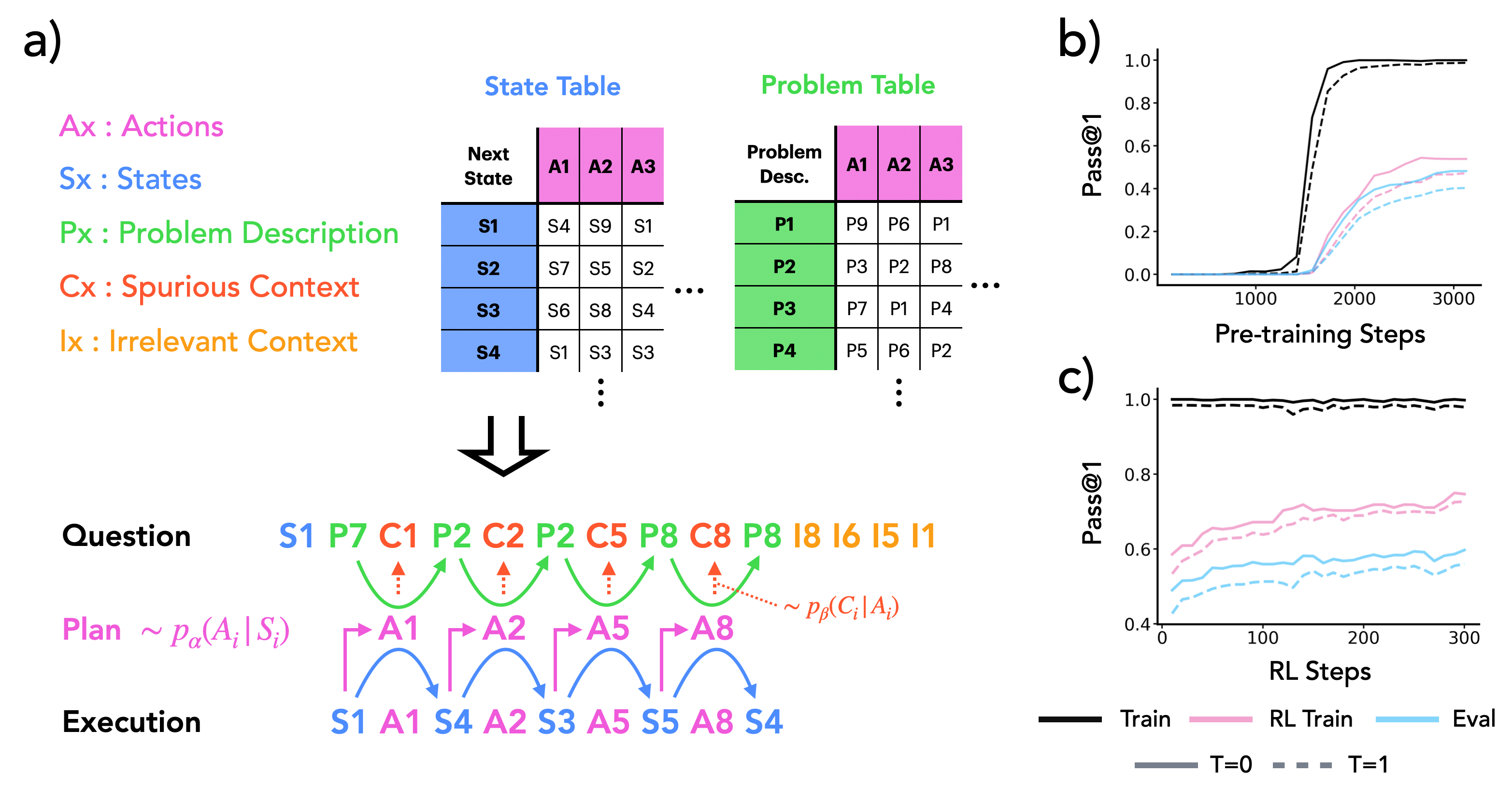}
    \vspace{-5pt}
    \caption{
    \textbf{Synthetic Math Setup and Training Curves a) Synthetic Setup.} Our setup is an analogy to mathematical problem solving. We design a graph navigation task where a model should internalize target states given the current state and action. However, the problem statement only encodes this action sequence (plan) indirectly and has spuriously correlated context which hint the action, which acts as a distractor.
    \textbf{b) Pre-training dynamics.} Dynamics of pass@1 accuracies during pretraining. The performance gap between the RL training set and the eval set is random and not systematic.
    \textbf{c) RL dynamics.} Dynamics of pass@1 accuracies during RL. We observe a generalization gap between the RL training set and the evaluation set.
    }\label{fig:syn_set_up}
\end{figure}

\subsection{Motivation}
Our empirical analysis of RLVR's impact on LLM mathematical problem solving offers a more nuanced understanding of the types of improvements RL delivers. However, the complexity of the training data of language models as well as the given math problems complicate the understanding of why and when we should expect the behavioral change observed in RLVR on MATH. This is especially important as there are clearly cases where RL reveals additional problems in both classical RL \cite{silver2017masteringchessshogiselfplay,schut2023bridginghumanaiknowledgegap} and RL in autoregressive transformers \cite{qin2025backtrackbacktracksequentialsearch}. Recently, researchers have developed simplified toy models and experimental setups that mirror key behaviors observed in large language models and other foundation models, enabling controlled perturbations and deeper understanding \cite{okawa2024compositionalabilitiesemergemultiplicatively,park2024emergencehiddencapabilitiesexploring,park2025competitiondynamicsshapealgorithmic,allenzhu2024physicslanguagemodels32,lubana2024percolationmodelemergenceanalyzing,jain2024makesbreakssafetyfinetuning,michaud2024quantizationmodelneuralscaling}. To this end, we develop a synthetic setup which emulates core aspects of mathematical problem solving in order to 1) reproduce and confirm our findings in a well-controlled setting and 2) determine what are the key conditions for RL to improve coverage.\looseness=-1

\subsection{Synthetic Setup}
The ideal synthetic setup should be a good analogy to the original setup in hand and reproduce core observed phenomena. Given these conditions are met, it is preferable to have the simplest setup possible. Here, we aim to capture these aspects of mathematical problem solving:
\vspace{-5px}
\begin{enumerate}[itemsep=1pt,labelindent=2pt,topsep=0pt,parsep=0pt,partopsep=0pt, align=left, leftmargin=*]
    \item There exists a ground truth action sequence to apply on the problem state, and one could deduce this action sequence from the problem statement in a learnable way.
    \item There exists a vast pretraining data with biases, spurious correlations and irrelevant context.
    \item The RL training and eval datasets, while not necessarily matching the pre-training distribution, are sampled i.i.d. from the same distribution—much like the questions in MATH \cite{Hendrycks2021-nv}.
\end{enumerate}
And we attempt to reproduce (See Sec.~\ref{sec:beyond_accuracy}):
\vspace{-5px}
\begin{enumerate}[itemsep=1pt,labelindent=2pt,topsep=0pt,parsep=0pt,partopsep=0pt, align=left, leftmargin=*]
    \item \textbf{Temperature Distillation:} RL reduces the performance gap between sampling at low temperature and high temperature.
    \item \textbf{Limited Coverage Increase:} RL shows a relatively modest coverage increase compared to Pass@1.
    \item \textbf{RL generalization gap:} RL demonstrates significant generalization gap between the training set reward and the eval set reward.
\end{enumerate}

The synthetic setup we designed is described in Fig.~\ref{fig:syn_set_up}. A ``math question" in this setup is defined as a set of actions to apply to an initial state where an action changes the state to a target state as described in the state table in Fig.~\ref{fig:syn_set_up}~a). The problem is solved by internalizing this state-action pair table, and applying them to the given initial state. However, the set of actions to apply is not directly given in the question statement, just like a math problem where the actions to be performed should be carefully deduced by thinking about the problem. Here, to make an analogy of this process, we encode the action sequence into sequence of ``problem states" which are also determined by a transition table. Thus, in order to solve the question correctly, the model must reconstruct the action sequence from the problem state transitions and apply them to the given state. In order to simulate spurious correlation, we generate a ``context" token for each sampled action which is highly correlated with the action number for the pre-training set, but not for the RL/eval set. Additionally, we add irrelevant context to the question part to allow memorization solutions as well. Overall, our synthetic data is designed to capture aspects of problem solving by having a more complex but accurate function to learn and spurious correlations to learn from. See Appendix ~\ref{appdx:synthetic_setup} for hyperparameters and further details about training, RL and eval data distributions as well as properties of the trained autoregressive model such as the choice of tokenizer or training hyperparameters.

\subsection{Main Results}
\paragraph{Training Dynamics.} Fig.~\ref{fig:syn_set_up}~b,c) shows the dynamics of pass@1 accuracy during pre-training and RL. During pre-training, we find that the RL training set and evaluation set both saturate at pass@1$\sim40\%$, while the model memorizes the whole training set. The performance between the RL training set and the eval set here is purely random as they are sampled i.i.d. These results suggests that the model indeed learned a superposition of the right problem solving strategy and one relying on spurious correlations, as discussed in Sec.~\ref{sec:execution_rubric}. In Fig.~\ref{fig:syn_set_up}~c), we can see that RL improves the model's performance on both splits unseen during pretraining, while there is a generalization gap between the RL training set performance and the evaluation set performance, as observed in real math experiments. Next, we decompose the performance gain of RL as we did in Sec.~\ref{sec:fine_grained}.

\begin{figure}
    \centering
    \includegraphics[width=0.85\linewidth]{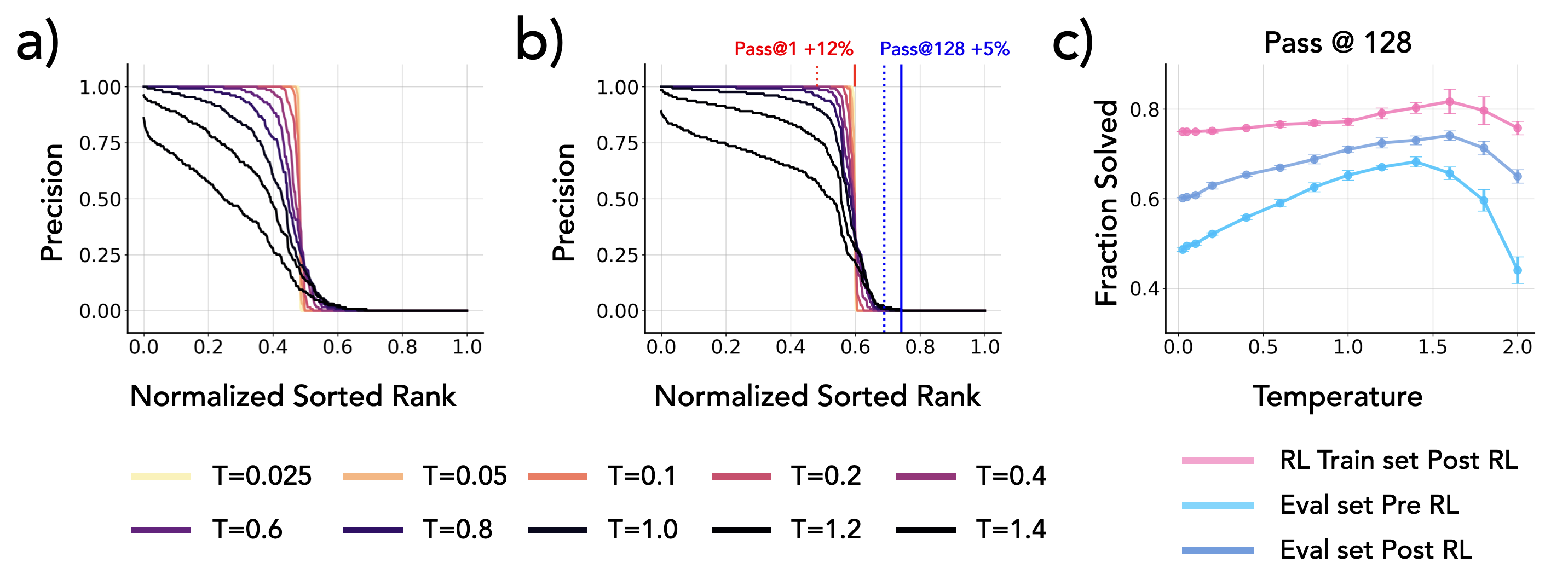}
    \vspace{-5pt}
    \caption{
    \textbf{RL improves Execution Robustness.} a, b) Precision of the trained autoregressive Transformer on 512 eval set questions before and after GRPO. c) Pass@128 of the RL training set and eval set depending on temperature.}
    \label{fig:syn_execution}
    \centering
    \includegraphics[width=0.85\linewidth]{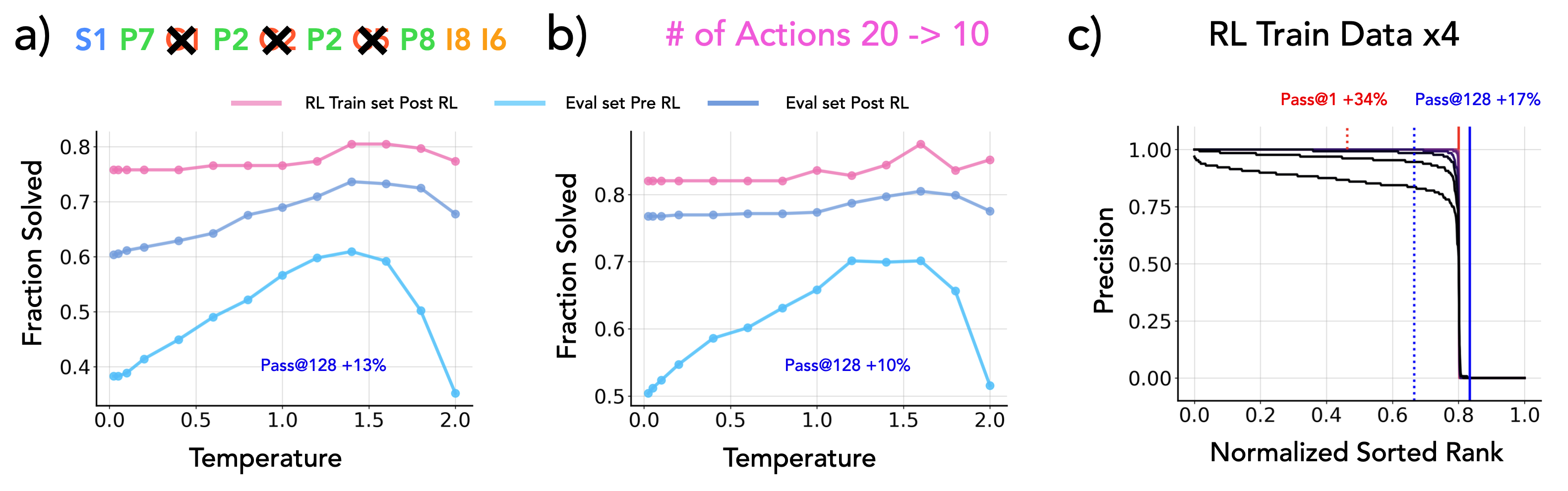}
    \caption{
    \textbf{When Does RL improves Coverage?} a) Pass@128 vs. Temperature, when the data has \textit{no context providing spurious correlations}. b) Pass@128 vs. Temperature, when the \textit{action space defining the problem is smaller}. c) Pass@128 vs. Temperature, when \textit{providing 4 times more RL data}.}\label{fig:syn_coverage}
\end{figure}

\paragraph{RL \textit{mostly} improves Execution Robustness.} Fig.~\ref{fig:syn_execution} breaks down the model's performance before and after GRPO similarly to Fig.~\ref{fig:qwen_temp}. Fig.~\ref{fig:syn_execution}~a,b) shows the precision over 512 unseen evaluation questions. \textit{We find that GRPO mostly robustifies precision when sampling at high temperature ($>0.5$), reproducing the Temperature Distillation effect seen in math.} In Fig.~\ref{fig:syn_execution}~c), we show the equivalent of Fig.~\ref{fig:qwen_temp}~(right) for our synthetic data. We find that while the is only a small coverage gain coming from GRPO. We discuss this small gain in Sec.~\ref{sec:conclusion}.

\paragraph{When RL improves Coverage.} Fig.~\ref{fig:syn_execution} hints that RL \textit{could} improve coverage. Using our synthetic model of problem solving, we can explore under what conditions the coverage boost by RL is significant, and understand when RL hits a wall. Fig.~\ref{fig:syn_coverage} shows three perturbations to the synthetic experiments. In Fig.~\ref{fig:syn_coverage}~a), We show that removing the context providing spurious correlations can boost the coverage gain given by RL. In Fig.~\ref{fig:syn_coverage}~b), we show that reducing the complexity/dimensionality of the problem by reducing the number of possible actions also allow both the coverage and greedy decoding accuracy to grow. Interestingly, this perturbation seem to primarily affect low temperature behavior. Finally, but perhaps most importantly, in Fig.~\ref{fig:syn_coverage}~c), we show that when we are in the regime where coverage \textit{is improving} (even slightly) we find that increasing the size(and diversity) of the RL training data greatly boosts the coverage increase. This is contrasting to recent findings which show that one can get most gains from very small amount of data \citep{wang2025reinforcementlearningreasoninglarge}. Our setup provides a contrasting case where performance \textit{does} scale with data, and provides a basis to investigate the cause of divergence of these regimes. This experiment also demonstrates that RL still somehow doesn't reach 100\% and his a wall, providing another thread to investigate.

\section{Conclusion and Discussion}
\label{sec:conclusion}
This work proposes a decomposition of mathematical reasoning in LLMs into three core capabilities: planning, execution, and verification. Using this framework, we conduct a in-depth analysis of reinforcement learning (specifically GRPO) and find that it primarily enhances execution robustness—making correct solutions more reproducible under sampling (a phenomenon we term temperature distillation)—but does not improve planning or expand the set of problems the model can solve (coverage). To further isolate this behavior, we design a minimal synthetic environment that mirrors our decomposition and replicates empirical observations. Crucially, this controlled setting reveals that RL can overcome the coverage wall when the model is exposed to actionable patterns that generalize across structurally similar problems.

\paragraph{Limitations and Future Work.}
Our study is limited to small-scale models and public datasets, and therefore cannot rule out the possibility of emergent behaviors in larger models trained with more extensive data. It is also plausible that curriculum design or expanded RL datasets could improve planning and coverage. Furthermore, while the synthetic environment reveals important mechanisms, its constrained state space makes it more susceptible to coverage improvements through brute-force sampling, which may not reflect the difficulty of generalization in real-world settings. Despite these limitations, our decomposition offers a conceptual tool for reasoning-focused analysis. We hope it encourages future work to move beyond aggregate accuracy and investigate how specific reasoning abilities emerge, interact, and improve under different training data and paradigms.


\section*{Acknowledgments}
TQ and DAM acknowledge support from the Kempner Institute, the Aramont Fellowship Fund, and the FAS Dean’s Competitive Fund for Promising Scholarship.
CFP and HT gratefully acknowledges the support of Aravinthan D.T. Samuel. CFP and HT are supported by NTT Research under the CBS-NTT Physics of Intelligence program. The computations in this paper were run on the FASRC cluster supported by the FAS Division of Science Research Computing Group at Harvard University. 
The authors thanks Zechen Zhang, Andrew Lee, Ekdeep Singh Lubana, Will Brown, Maya Okawa and Sara Fish for useful discussions.

\bibliographystyle{unsrtnat}
\bibliography{references,paperpile}

\clearpage
\tableofcontents

\clearpage
\appendix
\section{Training details}
\label{appx:experimental_details}
\subsection{Qwen2.5-1.5B Runs}
We list all important non-default hyperparameters used to train the Qwen-2.5-Instruct models (0.5B, 1.5B, and 7B; see Section~\ref{sec:beyond_accuracy}) in Table~\ref{tab:hyperparams}. The total number of training epochs was set to 60, with early stopping enabled based on convergence on the MATH test set. For all three model sizes, training converged within 100 steps.

\begin{table}[h]
    \centering
    \caption{Summary of hyperparameters used in our RL fine-tuning setup.}
    \vspace{10px}
    \label{tab:hyperparams}
        \begin{tabular}{@{}llr@{}}
        \toprule
        \textbf{Category} & \textbf{Hyperparameter} & \textbf{Value} \\
        \midrule
        \textbf{Data} 
        & max\_response\_length      & 4096 \\
        & train\_batch\_size         & 1024 \\
        \midrule
        \textbf{Actor} 
        & pop\_mini\_batch\_size     & 256 \\
        & use\_kl\_loss              & True \\
        & kl\_loss\_coef             & 0.001 \\
        & kl\_loss\_type             & low\_var\_kl \\
        \midrule
        \textbf{Optimizer} 
        & lr                         & 1e-6 \\
        & lr\_warmup\_steps\_ratio   & 0.0 \\
        & warmup\_style              & constant \\
        \midrule
        \textbf{Rollout} 
        & temperature               & 1.0 \\
        & response\_length          & 2048 \\
        & n                         & 8 \\
        \midrule
        \textbf{Trainer} 
        & total\_epochs             & 60 \\
        \bottomrule
        \end{tabular}
\end{table}
\vspace{10px}

\subsection{Computations Resources}
All models were trained using NVIDIA H100 80GB HBM3 GPUs. Qwen-2.5-0.5B-Instruct and Qwen-2.5-1.5B-Instruct were trained using 4 GPUs, while Qwen-2.5-7B-Instruct was trained using 8 GPUs. Each training run completed within 72 hours.

\section{Additional Evaluation Across Model, Scale, and Benchmark}
\label{appdx:additional_eval}
\begin{figure}[h]
    \centering
    \includegraphics[width=1.0\linewidth]{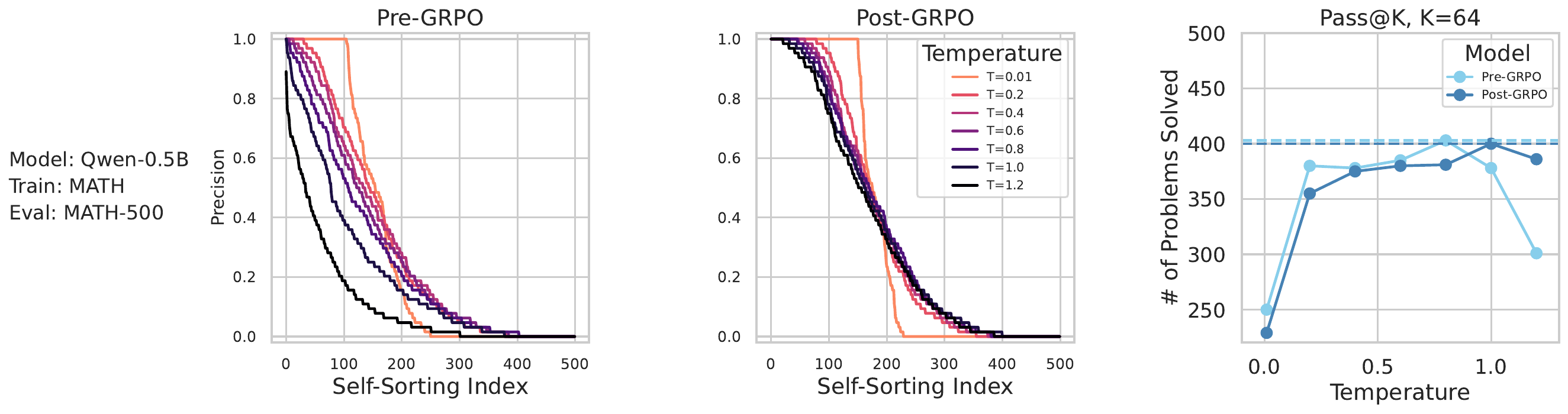}
    \includegraphics[width=1.0\linewidth]{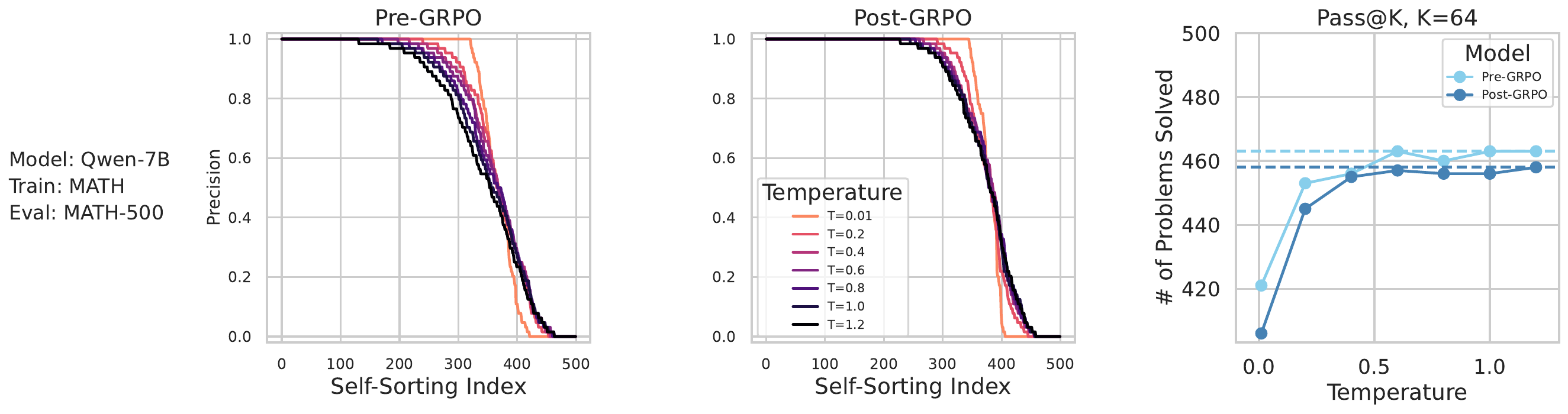}
    \includegraphics[width=1.0\linewidth]{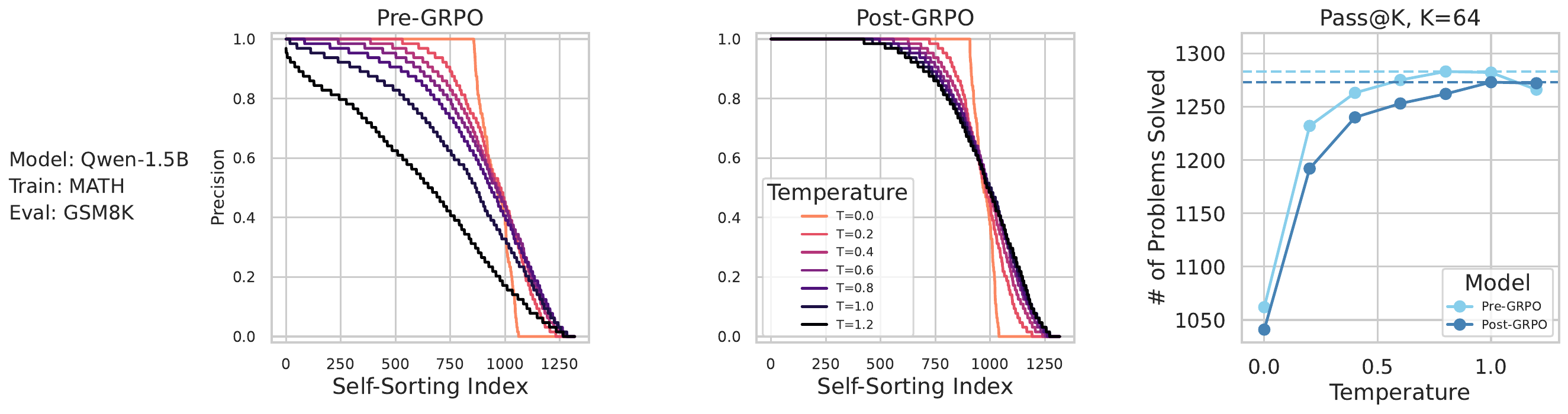}
    \includegraphics[width=1.0\linewidth]{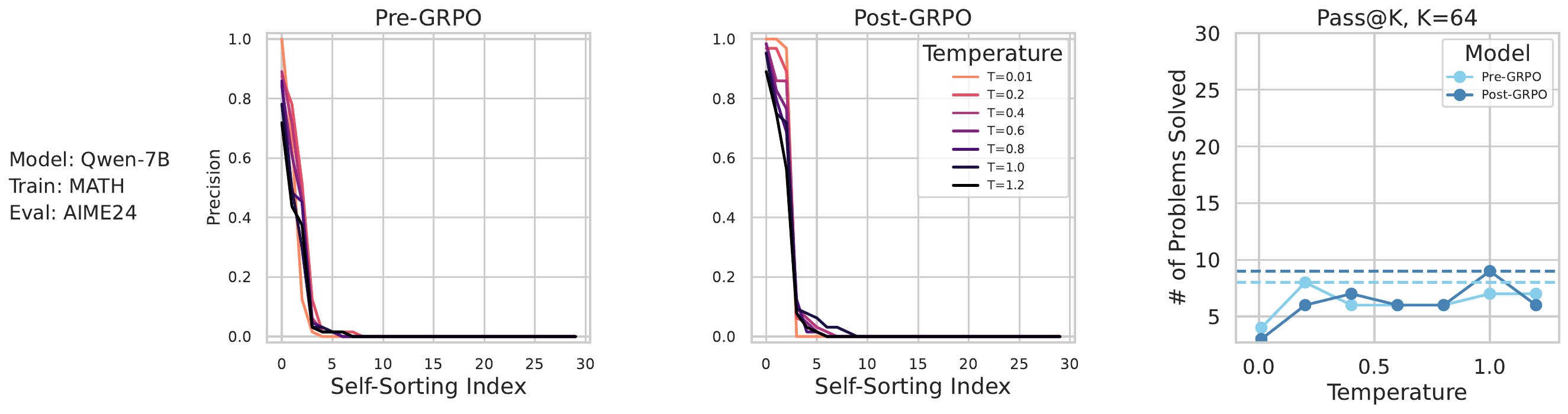}
    \includegraphics[width=1.0\linewidth]{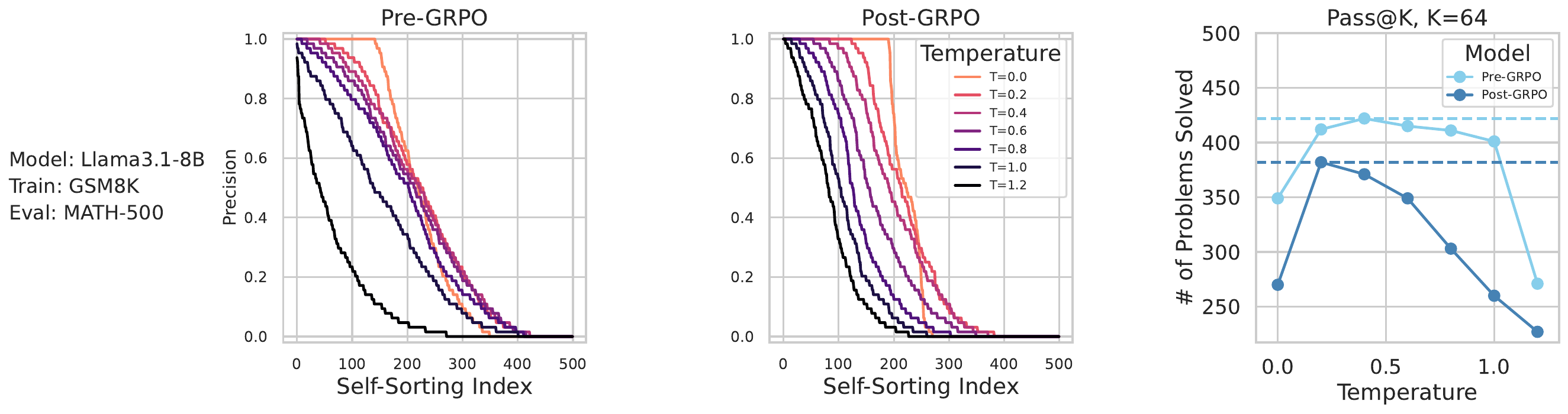}
    \caption{
    \textbf{The temperature distillation effect and coverage wall are consistent across settings.}
    We run GRPO across different model sizes, evaluation benchmarks, and model families, and observe the same qualitative outcomes: model precision becomes more stable ("sharpens") across sampling temperatures, while Pass@$k$ does not improve for large $k$, indicating the presence of a coverage wall.
}
    \label{fig:comprehensive_eval}
\end{figure}

To demonstrate that the temperature distillation effect (Sec.\ref{sec:temp_distill}) and the coverage wall (Sec.\ref{sec:coverage}) are universal phenomena, we reproduce Fig.\ref{fig:qwen_temp} across different model scales within the Qwen-2.5-Instruct family, as shown in Fig.\ref{fig:comprehensive_eval} (rows 1 and 2). We further show that these effects persist across different evaluation benchmarks (Fig.\ref{fig:comprehensive_eval}, rows 3 and 4), as well as in models from a different model family (Fig.\ref{fig:comprehensive_eval}, row 5). Concurrent work by \citet{Yue2025-se} provides additional complementary evaluations, confirming that the coverage wall effect also appears in coding tasks and with larger models.

\section{Justification for Coverage}
\label{appdx:coverage_k}
\begin{figure}[h]
    \centering
    \includegraphics[width=0.5\linewidth]{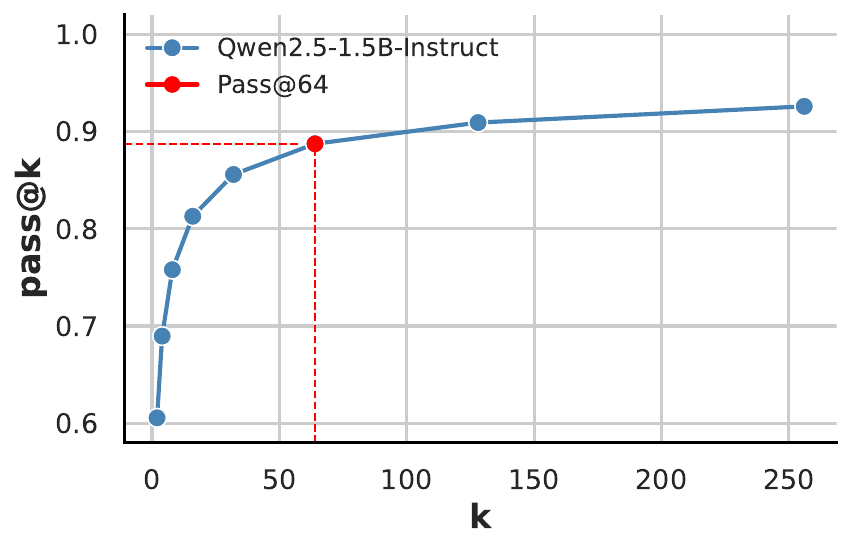}
    \caption{
    \textbf{Sweeping different $k$ values to measure coverage.} Sweeping through model's pass@$k$ on MATH-500, we conclude that $k=64$ is sufficiently large to measure how many problems are solvable by the model.
    }
    \label{fig:coverage_justify}
\end{figure}

We define \textit{coverage} as the model’s Pass@$k$ at an optimal sampling temperature for sufficiently large $k$ (Sec.\ref{sec:coverage}). Since Pass@$k$ increases monotonically with $k$, the choice of $k$ must balance practical compute constraints and the intended use case. \citet{Yue2025-se} show that as $k$ increases exponentially, models are able to solve more problems. In Fig.\ref{fig:coverage_justify}, we plot the Pass@$k$ performance of the Qwen2.5-1.5B-Instruct model on MATH-500 at $T = 0.6$, confirming the previously observed log-linear scaling behavior. Based on this visualization, we find that $k = 64$ is sufficiently large to reflect the model’s coverage within a reasonable compute budget.

\section{Additional Experiment on Qwen2.5-Instruct}
\begin{figure}[h]
    \centering
    \includegraphics[width=1.0\linewidth]{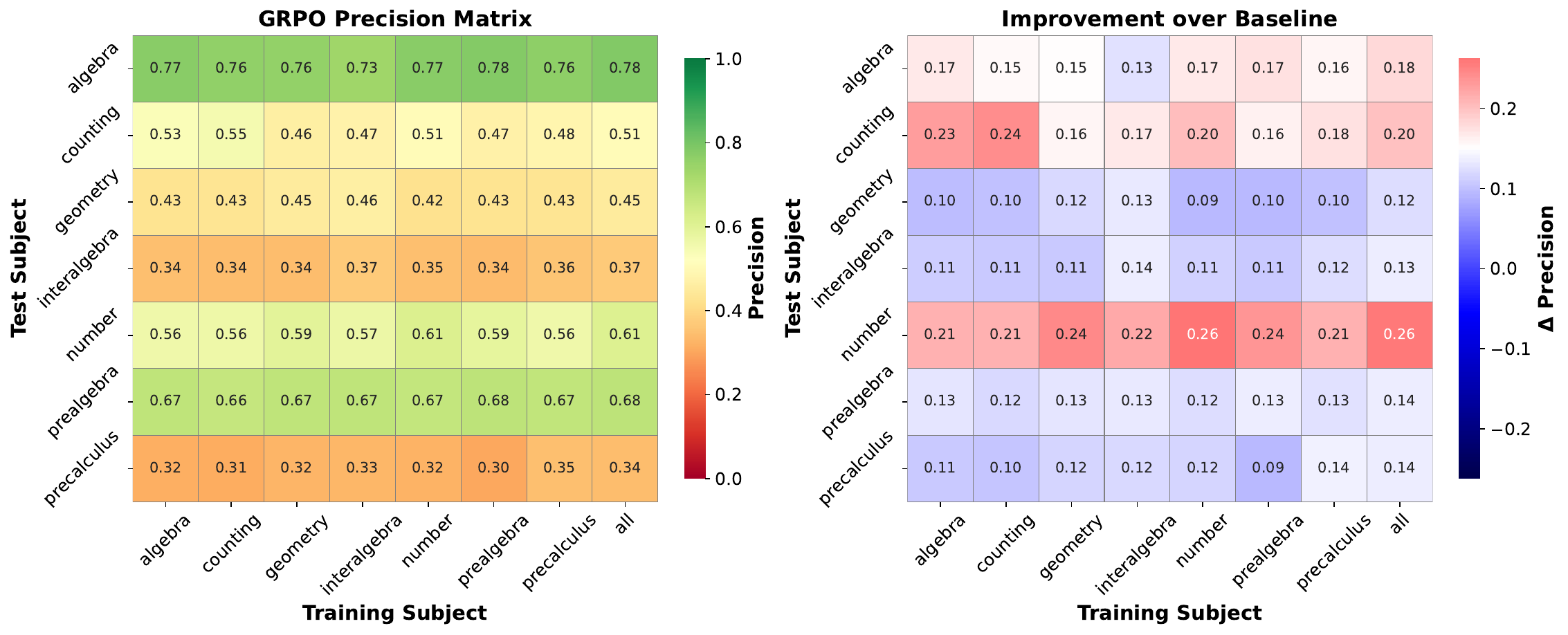}
    \caption{
    \textbf{GRPO improves mathematical reasoning in a subject-agnostic manner.} 
    We train GRPO models on seven subject-specific subsets of the MATH training data and evaluate their performance on number theory questions. Regardless of the training subject, all models show similar improvements on number theory, suggesting that GRPO does not yield subject-specific gains.
}
    \label{fig:qwen_subject}
\end{figure}

Figure \ref{fig:qwen_subject}.
In Sec.~\ref{sec:execution_rubric}, we argue that GRPO primarily addresses the spurious correlations models rely on when solving math problems. Using GPT-based annotation, we show that GRPO mainly reduces basic mathematical (i.e., elementary-level) and logical errors, but does not improve the model’s understanding of more advanced (i.e., high school–level) math. As an additional experiment, we conduct a per-subject analysis and find that GRPO fails to meaningfully enhance the model’s performance on high school math topics.

In this experiment, we partition the MATH training data by subject, using the subject categories provided by \citet{Hendrycks2021-nv}. Both the training and test sets cover the following seven subjects: algebra, counting, geometry, intermediate algebra, number theory, pre-algebra, and precalculus, each appearing in roughly equal proportions. We perform GRPO independently on each subject-specific training subset and evaluate performance on the corresponding subject in the MATH-500 test set.

In Fig.\ref{fig:qwen_subject} (\textit{left}), we report the post-GRPO model’s performance across different subjects, using average precision measured at $T = 1.0$. We observe that the values within each row are roughly uniform, indicating that regardless of the subject used for training, the model achieves similar performance across all test subjects. In Fig.\ref{fig:qwen_subject} (\textit{right}), we visualize the precision improvement over the pre-GRPO baseline. If the model had acquired subject-specific knowledge, we would expect a diagonal pattern in the heatmap, reflecting the greatest performance gains on test problems most closely related to the training subject. However, the heatmap appears nearly random, suggesting that GRPO does not lead to meaningful subject-specific improvements.

\clearpage
\section{Planning versus Execution}
\label{appdx:human_annot_example}
This section contains further details for experiments in Sec.~\ref{sec:direction_execution}.

\subsection{Prompt Details}
\begin{figure}[h]
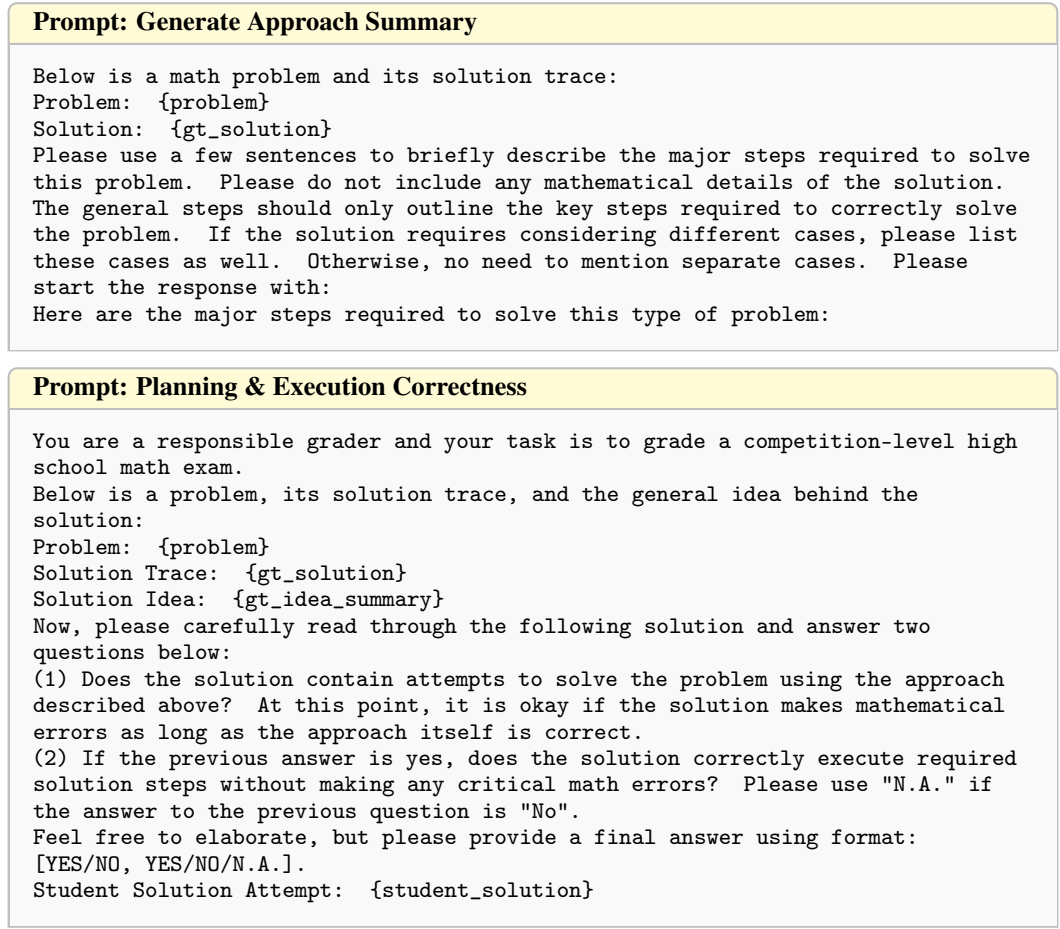

    \centering
    \begin{promptbox}[title={Prompt: Generate Approach Summary}]
Below is a math problem and its solution trace:

\textbf{Problem:} \texttt{\{problem\}}

\textbf{Solution:} \texttt{\{gt\_solution\}}

Please use a few sentences to briefly describe the major steps required to solve this problem.  
Please do not include any mathematical details of the solution.  
The general steps should only outline the key steps required to correctly solve the problem.  
If the solution requires considering different cases, please list these cases as well. Otherwise, no need to mention separate cases.  
Please start the response with:  

Here are the major steps required to solve this type of problem:
\end{promptbox}
\begin{promptbox}[title={Prompt: Planning \& Execution Correctness}]
You are a responsible grader and your task is to grade a competition-level high school math exam.  

Below is a problem, its solution trace, and the general idea behind the solution:  

\textbf{Problem:} \texttt{\{problem\}}  

\textbf{Solution Trace:} \texttt{\{gt\_solution\}}  

\textbf{Solution Idea:} \texttt{\{gt\_idea\_summary\}} 

Now, please carefully read through the following solution and answer two questions below:  

(1) Does the solution contain attempts to solve the problem using the approach described above? At this point, it is okay if the solution makes mathematical errors as long as the approach itself is correct.  

(2) If the previous answer is yes, does the solution correctly execute required solution steps without making any critical math errors? Please use "N.A." if the answer to the previous question is "No".  

Feel free to elaborate, but please provide a final answer using format: [YES/NO, YES/NO/N.A.].  

\textbf{Student Solution Attempt:} \texttt{\{student\_solution\}}
\end{promptbox}

    \caption{
    Prompts used for evaluating model reasoning. \textbf{Top:} Prompt used to generate the plan summary for a question from MATH-500. \textbf{Bottom:} Prompt used to grade whether the model’s solution contains the correct plan and executes it correctly.
}
    \label{fig:plan_exe_prompt}
\end{figure}

In Fig.~\ref{fig:plan_exe_prompt}, we show the exact prompts used to generate model solutions, as well as the prompts used to classify whether a solution contains the correct plan and whether it is executed correctly.

\subsection{Plan and Execution Failure Examples}
In Figs.\ref{fig:plan_eg_1}–\ref{fig:plan_eg_5}, we show five randomly selected examples where GPT identified that the Qwen2.5-1.5B-Instruct model failed to solve a problem due to an incorrect plan—i.e., not knowing the correct general approach. In Figs.\ref{fig:exec_eg_1}–\ref{fig:exec_eg_5}, we present five randomly selected examples where the model failed due to execution errors, despite having identified the correct plan.

\begin{figure}
    \centering
    \begin{llmbox}[title={Plan Failure Example 1}]
Let's analyze the student's solution attempt in light of the problem, the solution idea, and the official solution trace.

---

\#\#\# (1) Does the solution contain attempts to solve the problem using the approach described above?

**The approach described above involves:**

- Expressing sums and sums of squares in terms of \( e \) and the other variables.
- Applying an inequality (Cauchy-Schwarz) to relate the sum and sum of squares of the other variables.
- Simplifying to get a bound on \( e \).
- Checking equality conditions.

**Student's approach:**

- The student starts correctly by expressing \( a + b + c + d = 8 - e \).
- The student attempts to relate \( (a + b + c + d)^2 \) to \( a^2 + b^2 + c^2 + d^2 \) and the pairwise products \( S = ab + ac + \cdots \).
- The student tries to find \( S \) and then uses some algebraic manipulations to relate \( S \) and \( e \).
- However, the student does **not** explicitly apply Cauchy-Schwarz or any inequality to bound \( e \).
- Instead, the student tries to solve an equation involving \( S \) and \( e \) directly, which is not the approach described.

**Conclusion:** The student attempts to express sums and sums of squares in terms of \( e \) and the other variables, but does **not** apply the key inequality step (Cauchy-Schwarz) or any inequality to bound \( e \). The approach is more algebraic manipulation of sums and products rather than using the inequality approach.

**Answer to (1):** NO.

...

**Final answer:** [NO, N.A.]
\end{llmbox}
    \caption{Plan Failure Example 1}
    \label{fig:plan_eg_1}
\end{figure}

\begin{figure}
    \centering
  \begin{llmbox}[title={Plan Failure Example 2}]
  (1) Does the solution contain attempts to solve the problem using the approach described above?

The student attempts to solve the problem by simplifying the given rational expressions and testing a potential root by direct substitution. However, the student does not follow the key steps outlined in the solution idea, such as expanding the numerators and denominators, introducing a substitution to simplify the expressions, clearing denominators algebraically, or solving the resulting polynomial equation. Instead, the student jumps to testing a specific value \(x=2\) without algebraic manipulation or solving the equation systematically.

Therefore, the approach used by the student is not aligned with the described approach, which involves algebraic simplification and substitution rather than root testing.

Answer: NO

(2) If the previous answer is yes, does the solution correctly execute required solution steps without making any critical math errors?

Since the answer to (1) is NO, this question is not applicable.

Answer: N.A.

---

**Final answer:** [NO, N.A.]
\end{llmbox}

    \caption{Plan Failure Example 2}
    \label{fig:plan_eg_2}
\end{figure}

\begin{figure}
    \centering
    \begin{llmbox}[title={Plan Failure Example 3}]
(1) Does the solution contain attempts to solve the problem using the approach described above?

The approach described in the official solution involves rewriting all trigonometric expressions in terms of sine and cosine, combining terms over a common denominator, applying double angle and sum-to-product formulas, and simplifying step-by-step until a simple value is obtained.

The student's solution attempt starts correctly by expressing \(\tan 100^\circ\) as \(\frac{\sin 100^\circ}{\cos 100^\circ}\), which aligns with the first step of the approach. However, the subsequent steps diverge significantly from the described approach. The student attempts to use the double angle formula in a non-standard way and introduces multiplication by \(\tan 20^\circ\) without clear justification or connection to the original expression. The manipulation involving \(\frac{\tan 100^\circ + \cot 100^\circ}{\tan^2 100^\circ}\) and the final boxed expressions are unclear and do not follow from the initial expression logically.

Thus, while the student attempts to use trigonometric identities, the approach is not consistent with the systematic method described in the official solution idea.

**Answer to (1):** NO

(2) If the previous answer is yes, does the solution correctly execute required solution steps without making any critical math errors?

Since the answer to (1) is NO, this question is N.A.

---

**Final answer:** [NO, N.A.]

\end{llmbox}

    \caption{Plan Failure Example 3}
    \label{fig:plan_eg_3}
\end{figure}

\begin{figure}
    \centering
    \begin{llmbox}[title={Plan Failure Example 4}]
 Let's analyze the student's solution attempt in light of the problem and the provided solution idea.

---

(1) Does the solution contain attempts to solve the problem using the approach described above?

**Recall the approach:**

1. Visualize and describe the region formed by the union of the nonagon and the 1-unit neighborhood around its perimeter.
2. Analyze the boundary of this region, identifying straight and curved parts.
3. Calculate the lengths of these boundary segments.
4. Sum these lengths to find the total perimeter.

**Student's approach:**

- The student starts by discussing the area of the nonagon and then the area of a pentagon formed by points within 1 unit of the perimeter.
- The student attempts to calculate areas of polygons (nonagon and pentagon) using cotangent formulas.
- The student then claims the perimeter of \(S\) is the perimeter of the pentagon, which they calculate as \(5 \times 2 = 10\).

**Assessment:**

- The student focuses on areas rather than perimeter, which is the quantity asked for.
- The student introduces a pentagon, which is not relevant to the problem (the original polygon is a nonagon, and the neighborhood around its perimeter does not form a pentagon).
- The student does not analyze the boundary of \(S\) in terms of straight and curved segments.
- The student does not consider the rounding of corners or the circular arcs formed by the 1-unit neighborhood.
- The student does not sum lengths of different boundary parts.

**Conclusion:**

The student does not follow the approach described in the solution idea. They do not visualize or describe the boundary of \(S\) properly, nor do they analyze or calculate the perimeter by summing straight and curved segments.

---

(2) If the previous answer is yes, does the solution correctly execute required solution steps without making any critical math errors?

Since the answer to (1) is **No**, this question is **N.A.**

---

\#\#\# Final answer:
\[\boxed{\text{[NO, N.A.]}}\]
\end{llmbox}
  
    \caption{Plan Failure Example 4}
    \label{fig:plan_eg_4}
\end{figure}

\begin{figure}
    \centering
    \begin{llmbox}[title={Plan Failure Example 5}]
(1) Does the solution contain attempts to solve the problem using the approach described above?

The approach described in the solution idea involves:
- Expressing the distances from a point to each plane using the distance formula.
- Substituting these distances into the sum of squares equation.
- Simplifying to find the geometric locus.
- Recognizing the shape and calculating the volume.

The student's solution attempts to identify the locus as a sphere and calculate its volume. However, the student does not correctly use the distance formula for the planes. Instead, the student writes expressions like
\[ d_1^2 = \frac{(x - 0)^2 + (z - 0)^2}{1^2} = x^2 + z^2, \]
which is not the distance from a point to a plane but rather looks like a distance in the plane or some other incorrect expression. The student also misinterprets the planes and their distances, adding terms incorrectly.

So, while the student attempts to solve the problem by identifying the locus and calculating the volume, the key step of expressing the distances from the point to the planes using the correct distance formula is not done correctly.

Therefore, the student does not follow the approach described in the solution idea properly.

Answer to (1): NO.

(2) If the previous answer is yes, does the solution correctly execute required solution steps without making any critical math errors?

Since the answer to (1) is NO, the answer here is N.A.

---

Final answer: [NO, N.A.]
\end{llmbox}
    \caption{Plan Failure Example 5}
    \label{fig:plan_eg_5}
\end{figure}

\begin{figure}
    \centering
    \begin{llmbox}[title={Execution Failure Example 1}]
(1) Does the solution contain attempts to solve the problem using the approach described above?

Yes. The student correctly identifies the hourly wages, uses the exchange rates to convert the wages into US dollars, and compares the earnings to determine who earns the most. The student explicitly calculates total earnings for 8 hours first, then converts to USD, which is a valid approach consistent with the general idea.

(2) If yes, does the solution correctly execute required solution steps without making any critical math errors?

No. The student's calculations are mostly correct, but the final conclusion is incorrect. The student finds:

- Navin: 1280 / 32.35 \(\approx\) 39.72 USD  

- Luka: 200 / 5.18 \(\approx\) 38.53 USD  

- Ian: 272 / 6.95 \(\approx\) 39.59 USD  

From these, Navin earns the most (39.72 USD), followed by Ian (39.59 USD), then Luka (38.53 USD). However, the student concludes Ian earns the most, which contradicts their own calculations. This is a critical error in interpreting the results.

Final answer: [YES, NO]

\end{llmbox}

    \caption{Execution Failure Example 1}
    \label{fig:exec_eg_1}
\end{figure}

\begin{figure}
    \centering
    \begin{llmbox}[title={Execution Failure Example 2}]
  (1) Does the solution contain attempts to solve the problem using the approach described above?

Yes. The student attempts to analyze the sequence of turns and the probabilities associated with each turn, and they try to set up an equation for \(P_A\), the probability that Alice wins, which corresponds to step 6 in the solution idea (setting up equations for the probabilities and solving them). They also consider the probability of Alice winning on her first turn and the continuation of the game, which aligns with the general approach.

(2) If yes, does the solution correctly execute required solution steps without making any critical math errors?

No. The student makes a critical error in setting up the equation for \(P_A\). They write:
\[
P_A = \frac{1}{2} + \frac{1}{2} P_A,
\]
which implies that after Alice fails on her first turn (probability \(1/2\)), the probability that Alice eventually wins is again \(P_A\) multiplied by \(1/2\). However, this is incorrect because after Alice fails, Bob flips a coin, and only if Bob fails (with probability \(1/2\)) does the game return to Alice's turn with probability \(P_A\). The correct equation should be:
\[
P_A = \frac{1}{2} + \left(\frac{1}{2} \times \frac{1}{2}\right) P_A = \frac{1}{2} + \frac{1}{4} P_A.
\]
Solving this gives \(P_A = \frac{2}{3}\), not 1/2 or 1.

Thus, the student's final answer \(P_A = 1\) is incorrect, and the error is critical because it leads to a nonsensical probability of 1 (certainty) for Alice winning.

**Final answer:** [YES, NO]

\end{llmbox}

    \caption{Execution Failure Example 2}
    \label{fig:exec_eg_2}
\end{figure}

\begin{figure}
    \centering
    \begin{llmbox}[title={Execution Failure Example 3}]
  Let's analyze the student's solution attempt in light of the problem, the solution idea, and the solution trace.

---

**(1) Does the solution contain attempts to solve the problem using the approach described above?**

- The student attempts to convert the fraction \(\frac{137}{500}\) into its decimal form.
- They write the decimal expansion as \(0.274\), which is correct.
- They then try to identify the digits to the right of the decimal point and find the last nonzero digit.
- This aligns with the general approach described in the solution idea.

**Answer for (1):** YES

---

**(2) Does the solution correctly execute required solution steps without making any critical math errors?**

- The decimal conversion is correct: \(\frac{137}{500} = 0.274\).
- However, the student makes a confusing and incorrect statement in step 2: "the first two decimal places are 0.27" --- this is a misinterpretation since the decimal places are digits, not a decimal number.
- In step 3, the student says "the digits after the decimal point are 74," which is incorrect. The digits after the decimal point in \(0.274\) are 2, 7, and 4.
- In step 4, the student states "Since 7 is the only nonzero digit in the digits after the decimal point, the last nonzero digit is 4," which is contradictory and confusing.
- Despite the confusing wording, the final answer \(4\) is correct.
- The student's reasoning about the digits after the decimal point is flawed and inconsistent.
- The critical step is to identify the last nonzero digit to the right of the decimal point, which is the digit 4 in 0.274.
- The student's explanation is mathematically incorrect in parts but the final answer is correct.

**Answer for (2):** NO

---

\#\#\# Final grading summary:

**[YES, NO]**
\end{llmbox}
  
    \caption{Execution Failure Example 3}
    \label{fig:exec_eg_3}
\end{figure}

\begin{figure}
    \centering
    \begin{llmbox}[title={Execution Failure Example 4}]
(1) Does the solution contain attempts to solve the problem using the approach described above?

Yes. The student attempts to recognize the general form of the equation and compare it to standard conic forms. They discuss the transformation implied by the term \(\left(\frac{x}{2} - 3\right)^2\), identify shifts and scalings, and conclude the conic is an ellipse. This aligns with the approach steps: recognizing the form, rewriting/manipulating, identifying the conic, and confirming the identification.

(2) If yes, does the solution correctly execute required solution steps without making any critical math errors?

No. While the student correctly identifies the conic as an ellipse, the execution is somewhat vague and contains inaccuracies:

- The student states the equation "represents a rotated conic section," but the given equation is not rotated; it is shifted and scaled but aligned with the axes.
- The student writes the standard forms incorrectly for parabola and hyperbola (e.g., parabola is not expressed as a difference of squares equal to 1).
- The student does not explicitly rewrite the equation into a standard ellipse form with denominators, e.g., \(\frac{(x-6)^2}{4} + \frac{y^2}{10} = 1\), which would be a clearer and more rigorous step.
- The explanation about "scaling transformation on the y-coordinate" is unclear and not mathematically precise.
- The conclusion is correct, but the reasoning lacks rigor and clarity.

Final grading: [YES, NO]
\end{llmbox}
  
    \caption{Execution Failure Example 4}
    \label{fig:exec_eg_4}
\end{figure}

\begin{figure}
    \centering
    \begin{llmbox}[title={Execution Failure Example 5}]
\#\#\#(1) Does the solution contain attempts to solve the problem using the approach described above?

Yes. The student correctly identifies the need to rationalize the denominator, multiplies numerator and denominator by the conjugate, simplifies the denominator using difference of squares, expands the numerator, and attempts to express the result in the required form. This matches the general approach described.

\#\#\#(2) If yes, does the solution correctly execute required solution steps without making any critical math errors?

No. While the student correctly rationalizes the denominator and expands the numerator, the step where they try to rewrite the expression in the form \(A(1+\sqrt{B}) - (\sqrt{C} + \sqrt{D})\) is flawed. The student writes:

\[
2 + 2\sqrt{2} - \sqrt{3} - \sqrt{6} = 1 + 2\sqrt{2} + (2 - \sqrt{3} - \sqrt{6})
\]

which is incorrect because \(1 + 2\sqrt{2} + (2 - \sqrt{3} - \sqrt{6}) = 3 + 2\sqrt{2} - \sqrt{3} - \sqrt{6}\), not the original expression.

Then the student writes:

\[
1 + 2\sqrt{2} + 1 - \sqrt{3} - \sqrt{6} = (2\sqrt{2} + 1) - (\sqrt{3} + \sqrt{6})
\]

which is also incorrect because \(1 + 2\sqrt{2} + 1 - \sqrt{3} - \sqrt{6} = 2 + 2\sqrt{2} - \sqrt{3} - \sqrt{6}\), which is the original expression, but the grouping is not in the form \(A(1+\sqrt{B}) - (\sqrt{C} + \sqrt{D})\).

The correct way to write \(2 + 2\sqrt{2} - \sqrt{3} - \sqrt{6}\) in the form \(A(1+\sqrt{B}) - (\sqrt{C} + \sqrt{D})\) is:

\[
2 + 2\sqrt{2} - \sqrt{3} - \sqrt{6} = 2(1 + \sqrt{2}) - (\sqrt{3} + \sqrt{6})
\]

which directly identifies \(A=2\), \(B=2\), \(C=3\), and \(D=6\).

Thus, the student's final identification of constants and sum is correct, but the intermediate explanation and rearrangement are flawed.

Because the problem asks if the solution correctly executes the required steps without critical math errors, and the student's algebraic manipulation to express the form is incorrect (though the final answer is correct), this counts as a critical error in explanation.

**Final answer:**
[YES, NO]
\end{llmbox}
  
    \caption{Execution Failure Example 5}
    \label{fig:exec_eg_5}
\end{figure}
\clearpage
\section{Solution Tree Visualization}
\label{appdx:solution_tree}
This section contains details about the visualization tool used in Sec.~\ref{sec:execution_rubric}.

\subsection{Visualization Tool}
\begin{figure}[h]
    \centering
    \includegraphics[width=1.0\linewidth]{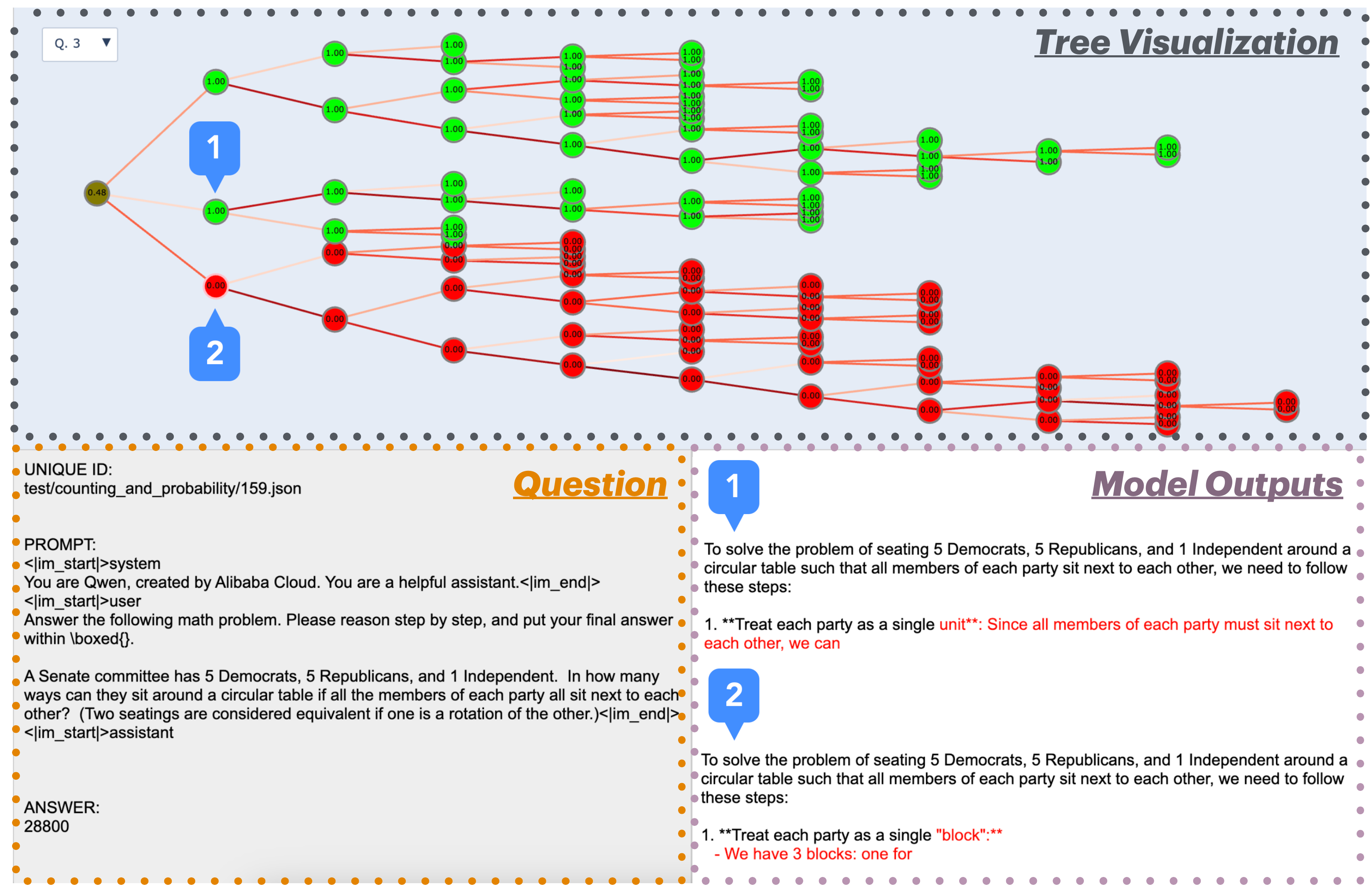}
    \caption{\textbf{A screenshot of the visualization tool we created to explore model's generation as a solution tree.}
    For a given problem, we sampled model's generations and construct a tree based on these generations. The tree branches out whenever the generations differ from each other. The color of the node indicating the accuracy when rolling out from that node and the numerical value indicating the likelihood of model reaching the node. }
    \label{fig:tree_ex1}
\end{figure}

In Fig.\ref{fig:tree_ex1}, we show a screenshot of the solution tree visualization tool used to explore model generations. The tool takes a set of model outputs for a given problem and constructs a tree, where branches are created whenever sampled solutions diverge. Each node is colored based on the accuracy of the solution when rolled out from that point, and labeled with the probability of reaching that node. An interactive panel displays the model’s output trace up to the currently selected node. The tree shown corresponds to the example in Fig.\ref{fig:exec_mistakes} (\textit{left}). An interactive \texttt{.html} version is included in \url{https://github.com/cfpark00/RL-Wall}, visualizing 64 sampled solutions for 10 questions.

\subsection{Further Examples}
In Fig.~\ref{fig:model_eg1}, we show two additional examples of model outputs, both of which are also included in the visualization tool examples. In the first example, the Qwen2.5-1.5B-Instruct model incorrectly assumes that an equilateral triangle has six sides instead of three. This leads to an incorrect final solution and reflects a basic factual error—one that an elementary school student would typically recognize. In the second example, the correct approach is to apply brute-force computation (e.g., calculating successive powers of 8) until a pattern is discovered. The model correctly begins this process but fails to recognize when the correct solution has been reached, continuing to compute indefinitely. This example highlights that, despite strong benchmark performance, the model still lacks basic logical reasoning skills.
\begin{figure}[h]
    \centering
    \begin{outputbopx}[title={Model Outputs Example 1}]
    \textbf{Question:}  
    
    A regular hexagon can be divided into six equilateral triangles. If the perimeter of one of the triangles is 21 inches, what is the perimeter, in inches, of the regular hexagon?

    \textbf{Output:}
    
    To determine the perimeter of the regular hexagon, we start by understanding the relationship between the hexagon and the equilateral triangles it is composed of. A regular hexagon can be divided into six equilateral triangles, each with the same side length.
    
    Given that the perimeter of one of these equilateral triangles is 21 inches, we can find the length of one side of the equilateral triangle. Since all sides of an equilateral triangle are equal, the length of one side is:
    \textcolor{red}{
    \[
    \text{Side length of the equilateral triangle} = \frac{21}{6} = 3.5 \text{ inches}
    \]
    }
    
    Next, we need to find the perimeter of the regular hexagon. A regular hexagon has six sides, and each side is equal to the side length of the equilateral triangles. Therefore, the perimeter of the hexagon is:
    \[
    \text{Perimeter of the hexagon} = 6 \times 3.5 = 21 \text{ inches}
    \]
    
    Thus, the perimeter of the regular hexagon is:
    \[
    \boxed{21}
    \]
    \end{outputbopx}
        \begin{outputbopx}[title={Model Outputs Example 2}]
    \textbf{Question:}  
    
    Find the units digit of $18^6.$

    \textbf{Output:}
    
    To find the units digit of \(18^6\), we need to determine the units digit of the number \(18\) raised to the power of \(6\). The units digit of a number is the same as the units digit of its last digit when the number is written in its decimal form. Therefore, we only need to consider the units digit of \(8\) when raised to the power of \(6\).

Let's break it down step by step:

1. Identify the units digit of \(8\):
   The units digit of \(8\) is \(8\).

2. Determine the pattern in the units digits of powers of \(8\):
   - \(8^1 = 8\) (units digit is \(8\))
   
   - \(8^2 = 64\) (units digit is \(4\))
   
   - \(8^3 = 512\) (units digit is \(2\))
   
   - \(8^4 = 4096\) (units digit is \(6\))
   
   - \(8^5 = 32768\) (units digit is \(8\))
   
   - \(8^6 = 262144\) (units digit is \(4\))

   \textcolor{red}{
   - \(8^7 = 2097152\) (units digit is \(2\))}
   
   \textcolor{red}{
   - \(8^8 = 16777216\) (units digit is \(6\))}
   
   \textcolor{red}{
   - \(8^9 = 134217728\) (units digit is \(8\))}
   
   \textcolor{red}{
   - \(8^{10} = 1073741824\) (units digit is \(4\))}
   
   \textcolor{red}{
   - \(8^{11} = 8589934592\) (units digit is \(2\))}
   
   \textcolor{red}{
   - \(8^{12} = 703687441776\) (units digit is \(6\))}
   
   \textcolor{red}{
   - \(8^{13} = 5629499534216\) (units digit is \(8\))}
   
   \textcolor{red}{
   - \(8^{14} = 45035996273704\) (units digit is \(4\))}
   
   \textcolor{red}{
   - \(8^{15} = 3602879701896384\) (units digit is \(2\))}
   
   \textcolor{red}{
   - \(8^{16} = 288230376151711760\) (units digit is \(6\))}
   
   \textcolor{red}{
   - \(8^{17} = 2305843009213693952\) (units digit is \(8\))}
   
   \textcolor{red}{
   - \(8^{18} = 184756814741950406400\) (units digit is \(4\))}
   
   \textcolor{red}{
   - \(8^{19} = 1478583885935203251200\) (units digit is \(2\))}

   $\vdots$
   
    \end{outputbopx}
    
    \caption{Examples of Model Generated Solutions on MATH-500 Questions}
    \label{fig:model_eg1}
\end{figure}

\clearpage
\section{Execution Error Annotations}
\label{appdx:rubric_details}
This section contains additional details for the execution grading done in Sec.~\ref{sec:execution_rubric}. 

\subsection{Prompts}
In Fig.~\ref{fig:exec_rubric_prompt}, we list the prompt used to classify the type of execution errors. 

\begin{figure}[h]
    \centering
    \begin{promptbox}[title={Prompt: Factual and Logic Execution Errors}]
You are grading a high school competition exam. Below is the grading rubric. I will provide you the question and the student's solution attempt. Please make your best judgments on the three grading criteria below. Feel free to elaborate, but please output your final grading in the format of \texttt{\textbackslash boxed\{YES/NO, ELEMENTARY/HIGH/NA, YES/NO\}}. It is very important that you follow the exact format in verbatim for the final grade. Please do not add any additional formatting in your final answer.

\textbf{Grading Rubric:}
(1). Does the solution contain basic mathematical factual mistakes? Here we are talking about simple plain mathematical facts that are in direct contradiction to well-known mathematical knowledge. Examples of mathematical facts are: how many sides does a triangle have, is 2 an odd number. Note that if the student makes a computational or algebraic mistake when manipulating mathematical operations, we DO NOT count this as a factual mistake. We are looking for mistakes that can be spotted even without the problem or solution context. In other words, the mistake is obvious in isolation.

(2). If the answer to (1) is yes, is the math factual mistake an elementary (including middle school) one or a high-school one? Please output \texttt{ELEMENTARY} or \texttt{HIGH}. If the answer to (1) is no, please output \texttt{NA}.

(3). Does the solution contain basic logic mistakes (no math involved to spot such an error) that are obviously nonsensical? We are talking about very basic logic mistakes that even a human with no mathematical knowledge can identify. To reiterate, we are looking for plain simple logic errors. For example, the student self-contradicting what was said before, or making a conclusion without clear logic from existing steps. Again, mistakes in complex math concepts do not count here.

\textbf{Problem:} \texttt{\{problem\}}

\textbf{Solution Attempt:} \texttt{\{student\_solution\}}
\end{promptbox}
    \caption{Execution Failure Annotation Prompts}
    \label{fig:exec_rubric_prompt}
\end{figure}

\subsection{Example}
We list 10 randomly selected examples below. Fig~\ref{fig:exec_rubric_eg_1}, \ref{fig:exec_rubric_eg_2}, \ref{fig:exec_rubric_eg_3}, \ref{fig:exec_rubric_eg_4}, \ref{fig:exec_rubric_eg_5} show elementary and basic logic errors and Fig~\ref{fig:exec_rubric_eg_6}, \ref{fig:exec_rubric_eg_7}, \ref{fig:exec_rubric_eg_8}, \ref{fig:exec_rubric_eg_9}, \ref{fig:exec_rubric_eg_10} show examples of advanced logic errors. 

\begin{figure}
    \centering
    \begin{llmbox2}[title={Example: Elementary \& Logic Error Example 1}]
  (1) The solution contains several factual mistakes regarding modular arithmetic and divisibility conditions. For example, the student incorrectly states that for \(100 - 4n \equiv 1 \pmod{3}\), it implies \(n \equiv 0 \pmod{3}\), which is incorrect. Since \(4 \equiv 1 \pmod{3}\), the congruence \(100 - 4n \equiv 1 \pmod{3}\) simplifies to \(1 - n \equiv 1 \pmod{3}\), so \(n \equiv 0 \pmod{3}\) is correct, but the student then continues with inconsistent and incorrect modular reasoning for \(k\) and \(j\). Also, the student repeatedly tests values of \(j\) and \(k\) but miscalculates or misinterprets the divisibility conditions, leading to contradictions and incorrect conclusions about integer values of \(m\). The final boxed answer 26 is not justified by the work shown and contradicts the earlier partial results.

(2) These mistakes are elementary modular arithmetic errors, which are typically taught in middle or early high school.

(3) The solution also contains basic logic mistakes: the student contradicts themselves by stating certain values are not integers, then later using those values to conclude a final answer. The testing of values is inconsistent and the conclusion is not logically supported by the steps shown.

\boxed{YES, ELEMENTARY, YES}
    \end{llmbox2}
    \caption{Basic Math and Logic Error Example 1}
    \label{fig:exec_rubric_eg_1}
\end{figure}

\begin{figure}
    \centering
    \begin{llmbox2}[title={Example: Elementary \& Logic Error Example 2}]
    Step-by-step analysis:

1. The prime numbers less than 30 are correctly listed as:
   2, 3, 5, 7, 11, 13, 17, 19, 23, 29

2. The odd numbers greater than zero are correctly listed as:
   1, 3, 5, 7, 9, 11, 13, 15, 17, 19, 21, 23, 25, 27

3. The intersection should be the prime numbers less than 30 that are also odd numbers greater than zero. Since 2 is prime but not odd, it should not be in the intersection.

4. The student incorrectly includes 2 in the intersection set, which is a factual mistake about the parity of 2 (2 is even, not odd).

5. The student concludes the intersection has 10 elements, but excluding 2, the intersection has 9 elements:
   3, 5, 7, 11, 13, 17, 19, 23, 29

6. This is a basic mathematical fact error (2 is not odd).

7. There is no basic logic mistake; the student's reasoning is consistent except for the factual error about 2.

Final grading:

(1) Basic mathematical factual mistake? YES (2 is not odd but included in intersection)  
(2) Is the mistake elementary or high school? ELEMENTARY (parity of 2 is elementary knowledge)  
(3) Basic logic mistake? NO

\boxed{YES, ELEMENTARY, NO}
    \end{llmbox2}
    \caption{Basic Math and Logic Error Example 2}
    \label{fig:exec_rubric_eg_2}
\end{figure}

\begin{figure}
    \centering
    \begin{llmbox2}[title={Example: Elementary \& Logic Error Example 3}]
    (1) Basic mathematical factual mistakes: The student correctly identifies amplitude, period, and vertical shift. The student also correctly uses the sine function properties and the general form. The only questionable part is the phase shift calculation: the student initially says the phase shift is \(-\frac{\pi}{2}\), but then solves the equation \(\sin(-\frac{3\pi}{2} + C) = 1\) and finds \(C = \pi + 2k\pi\), concluding the smallest \(C\) is \(\pi\). Then the student contradicts this by saying the phase shift is \(-\frac{\pi}{2}\). This is a logical inconsistency but not a factual mathematical mistake. The sine function's maximum occurs at \(\frac{\pi}{2}\) plus multiples of \(2\pi\), so the student's equation and solution for \(C\) is correct. The confusion is in interpreting the phase shift and the smallest possible \(c\). There is no outright factual mathematical error like "2 is odd" or "a triangle has 4 sides."

(2) Since no factual mathematical mistake is found, this is NA.

(3) Basic logic mistakes: The student contradicts themselves about the phase shift value. They first say the phase shift is \(-\frac{\pi}{2}\), then solve for \(C\) and get \(\pi\), then say the phase shift is \(-\frac{\pi}{2}\) again. This is a basic logic inconsistency: the student does not clearly justify why the smallest possible \(c\) is \(-\frac{\pi}{2}\) after finding \(C = \pi + 2k\pi\). The conclusion is not logically consistent with the steps shown.

Final grading: \boxed{NO, NA, YES}
    \end{llmbox2}
    \caption{Basic Math and Logic Error Example 3}
    \label{fig:exec_rubric_eg_3}
\end{figure}

\begin{figure}
    \centering
    \begin{llmbox2}[title={Example: Elementary \& Logic Error Example 4}]
    (1) The solution contains multiple factual mistakes:
- The student claims that the denominator of the reciprocal (which is an odd integer) must be of the form \(2^m\) to have a terminating decimal. This is incorrect because \(2^m\) is always even, and the problem restricts to odd integers less than 2010. So no odd integer except 1 can be a power of 2.
- The student also incorrectly counts the number of odd integers less than 2010 as 1004 (which is correct), but then counts 1001 values of \(2k+1\) as powers of 2 less than 2010, which is impossible.
- The student states the probability as \(\frac{1001}{505} = \frac{1}{5}\), which is mathematically incorrect since 1001/505 is not equal to 1/5.
- The final sum \(a+b=506\) is inconsistent with the previous statements.

(2) These mistakes are elementary because they involve misunderstanding of basic properties of odd numbers and powers of 2, and simple arithmetic errors.

(3) The solution also contains basic logic mistakes:
- The student claims that the denominator must be a power of 2, which contradicts the fact that the denominator is odd.
- The student's counting and probability calculation are inconsistent and nonsensical.
- The conclusion that the probability is \(\frac{1}{5}\) and \(a+b=506\) is not logically supported by the steps.

\boxed{YES, ELEMENTARY, YES}
    \end{llmbox2}
    \caption{Basic Math and Logic Error Example 4}
    \label{fig:exec_rubric_eg_4}
\end{figure}

\begin{figure}
    \centering
    \begin{llmbox2}[title={Example: Elementary \& Logic Error Example 5}]
    (1) Basic mathematical factual mistakes: The student correctly computes the initial terms and applies the recurrence relation without contradicting any basic mathematical facts. There is no obvious factual error such as miscounting sides of a polygon or misclassifying numbers. The computations and definitions are consistent with standard math facts.

(2) Since there is no factual mistake, this is NA.

(3) Basic logic mistakes: The solution attempt shows some unclear and confusing reasoning, especially in the middle where the student tries to find a pattern in sums of four terms but then writes an expression for \(x_{13}\) in a convoluted way. The student also repeats the expression \(x_{531} + x_{753} + x_{975}\) multiple times without explanation and then jumps to a conclusion that the sum is 138 without clear logical justification or demonstration of periodicity or how these indices relate to the sums computed. This is a basic logic flaw: the conclusion is not supported by the steps shown, and the reasoning is not coherent or complete. The student fails to logically connect the problem's requirements to the final answer.

Final grading: \boxed{NO, NA, YES}
    \end{llmbox2}
    \caption{Basic Math and Logic Error Example 5}
    \label{fig:exec_rubric_eg_5}
\end{figure}

\begin{figure}
    \centering
    \begin{llmbox2}[title={Example: High School Math Error Example 1}]
    Problem: 
 The Fibonacci sequence is the sequence 1, 1, 2, 3, 5, $\ldots$ where each term is the sum of the previous two terms. What is the remainder when the $100^{\mathrm{th}}$ term of the sequence is divided by 4? 
 Solution Attempt: 
 To find the remainder when the \( 100^{\mathrm{th}} \) term of the Fibonacci sequence is divided by 4, we start by examining the sequence modulo 4. The Fibonacci sequence modulo 4 is periodic, and we can determine its length by calculating the sequence's terms modulo 4 until we detect repetition.

The initial terms of the Fibonacci sequence are:
\[
1, 1, 2, 3, 0, 3, 3, 2, 1, 0, 1, 1, 2, 3, 1, 0, ...
\]

By writing these terms in groups of four, we get a repeating pattern:
\[
(1, 1, 2, 3)(0, 3, 3, 2)(1, 0, 1, 1)(2, 1, 0, 1)(1, 1, 2, 3)(0, 3, 3, 2)\ldots
\]
From this observation, it is evident that the sequence repeats every 6 terms. Therefore, the sequence is periodic with period 6, and the \( 100^{\mathrm{th}} \) term corresponds to the same position as the \( \text{100 mod 6}^{\mathrm{th}} = 4^{\mathrm{th}} \) term in the repeating block.

Looking at the sequence modulo 4 starting from the initial terms:
\[
1, 1, 2, 3, 0, 3, 3, 2, 1, 0, 1, 1, 2, 3, 1, 0, ...
\]
The \( 4^{\mathrm{th}} \) term in this block is 3. Therefore,
\[
\boxed{3}
\]
 
--------------------

Response: 

  (1) The student claims the Fibonacci sequence modulo 4 has period 6, but the initial terms modulo 4 they list do not support this. The sequence modulo 4 actually has a period of 12, not 6. The student's grouping into blocks of four terms and then concluding a period of 6 is incorrect. This is a factual mistake about the periodicity of the Fibonacci sequence modulo 4, which is a well-known fact.

(2) This mistake about the period length of the Fibonacci sequence modulo 4 is a high-school level fact, as it involves understanding modular arithmetic and periodicity in sequences, typically taught in high school.

(3) The logic of using periodicity to find the remainder is sound, and the student does not contradict themselves or make nonsensical conclusions. The error is factual, not logical.

Final grading: \boxed{YES, HIGH, NO}
    \end{llmbox2}
    \caption{High School Level Math Error Example 1}
    \label{fig:exec_rubric_eg_6}
\end{figure}

\begin{figure}
    \centering
    \begin{llmbox2}[title={Example: High School Math Error Example 2}]
    Step 3 contains an error in the calculation of the cross product. The cross product of \(\mathbf{a} = (1,1,1)\) and \(\mathbf{v} = (x,y,z)\) should be:

\[
\mathbf{a} \times \mathbf{v} = \begin{pmatrix} 1 \cdot z - 1 \cdot y \\ 1 \cdot x - 1 \cdot z \\ 1 \cdot y - 1 \cdot x \end{pmatrix} = \begin{pmatrix} z - y \\ x - z \\ y - x \end{pmatrix}.
\]

However, the student wrote the second component as \(z - x\) instead of \(x - z\). This is a factual mistake in the cross product formula.

Because the cross product formula is a basic vector operation taught in high school, this is a high school level factual mistake.

The rest of the solution proceeds logically from the incorrect cross product expression, so no basic logic mistakes are present.

\boxed{YES, HIGH, NO}
    \end{llmbox2}
    \caption{High School Level Math Error Example 2}
    \label{fig:exec_rubric_eg_7}
\end{figure}

\begin{figure}
    \centering
    \begin{llmbox2}[title={Example: High School Math Error Example 3}]
    (1) The student states that there are 4 sets of points that lie on the same straight line, but then counts 3 rows + 3 columns = 6 sets, divides by 2 to get 3 sets, and adds 2 diagonals to get 5 sets total. This is incorrect. The rows and columns are distinct lines, so there are 3 rows + 3 columns = 6 lines, not 3. Dividing by 2 is a mistake because rows and columns are different lines and should not be double counted. Also, the student misses other lines with 3 points, such as the lines with slope 1 or -1 that pass through other points (e.g., the middle points in the grid). So the count of lines with 3 points is incorrect. This is a factual mistake about counting lines in a grid, which is a well-known fact in combinatorics and geometry.

(2) This mistake is about counting lines in a 3x3 grid, which is a standard high school level combinatorics/geometry problem. So the factual mistake is HIGH level.

(3) The logic of the solution is consistent with the student's incorrect counting method. There is no self-contradiction or nonsensical logic beyond the counting error. So no basic logic mistake.

Final grading: \boxed{YES, HIGH, NO}
    \end{llmbox2}
    \caption{High School Level Math Error Example 3}
    \label{fig:exec_rubric_eg_8}
\end{figure}

\begin{figure}
    \centering
    \begin{llmbox2}[title={Example: High School Math Error Example 4}]
    (1) The solution contains a factual mistake in modular arithmetic and divisibility reasoning. Specifically, the student incorrectly concludes that \( m = \frac{4}{3}(26 - 3k) \) must be an integer, and then tries to impose further divisibility conditions on \( k \) that are not consistent. The step where the student tries to solve for \( m \) in terms of \( m \) again (step 5) is nonsensical and shows a misunderstanding of the variables and substitution. This is a factual error in modular arithmetic and divisibility, which is a high school level topic.

(2) Since the mistake involves modular arithmetic and divisibility conditions, which are typically taught in high school, this is a HIGH level factual mistake.

(3) The solution does not contain basic logic mistakes that are obviously non-sensical from a purely logical standpoint. The student attempts a logical approach, even if flawed mathematically.

Final grading: \boxed{YES, HIGH, NO}
    \end{llmbox2}
    \caption{High School Level Math Error Example 4}
    \label{fig:exec_rubric_eg_9}
\end{figure}

\begin{figure}
    \centering
    \begin{llmbox2}[title={Example: High School Math Error Example 5}]
    (1) The solution contains several factual mistakes:

- Step 8 claims that because the coefficients are symmetric, the other root will be the conjugate of 4, i.e., \(x = -4\). This is incorrect. Symmetry of coefficients does not imply roots come in pairs \(r\) and \(-r\). This is a misunderstanding of polynomial root properties.

- The evaluation of the polynomial at \(x = -4\) is incorrect:
\[
(-4)^4 - 16(-4)^2 + 5(-4) - 20 = 256 - 256 - 20 - 20 = -40 \neq -60
\]
The student wrote \(-60\), which is a computational mistake, but computational mistakes are not counted as factual mistakes.

- The student only tests a few roots and concludes \(x=4\) is the only root, which is logically incomplete.

- The student substitutes \(y = \sqrt{-5x + 44}\) only, ignoring the negative root \(y = -\sqrt{-5x + 44}\) in the substitution step, which is a conceptual error.

- The student finds only one root \(x=4\) and corresponding \(y=2\sqrt{6}\), and then incorrectly assigns \(y=0\) for \(x=-4\) without verifying if \((x,y)\) satisfies both equations.

- The final product calculation uses \(y=0\) from \(x=-4\), but since \((x,y) = (-4,0)\) is not a solution (the polynomial does not vanish at \(x=-4\)), this is invalid.

These errors show misunderstanding of polynomial root properties and solution verification, which are high school level concepts.

(2) The mistakes are high school level.

(3) The solution contains basic logic mistakes:

- The student assumes the root at \(-4\) without proper justification.

- The student ignores the negative root branch of the square root in substitution.

- The student concludes the product of \(y\)-coordinates is zero by including a non-solution point.

These are basic logical errors that make the solution non-sensical.

Final grading:

\boxed{YES, HIGH, YES}
    \end{llmbox2}
    \caption{High School Level Math Error Example 5}
    \label{fig:exec_rubric_eg_10}
\end{figure}
\clearpage
\section{Details on Synthetic Setup}
\label{appdx:synthetic_setup}
This section contains further details about the synthetic setups studied in Section \ref{sec:syn_math}.

\subsection{Model}
\label{appdx:synthetic_setup_model}

We use a standard autoregressive Transformer \citep{vaswani2023attentionneed} model. In specific, we use the \texttt{Qwen-2} architecture\citep{Yang2024-db}. Table~\ref{tab:toy_qwen2_arch} show the hyperparameters of the Transformer model. The vocabulary space is 29, constituting of 20 number tokens (0$\sim$19), required but never used ``unknown" token <unk>, beginning of sequence token <bos>, padding token <pad>, prompt/completion separation token ``:", and S, P, C, A, I tokens defined in Sec.~\ref{appdx:synthetic_setup}.

\begin{table}[ht]
\centering
\caption{Hyperparameters of the Qwen2 Transformer model for synthetic data training.}\label{tab:toy_qwen2_arch}
\vspace{10pt}
\begin{tabular}{ll}
\toprule
\textbf{Hyperparameter} & \textbf{Value} \\
\midrule
Vocabulary size & 29 (20 numbers, <unk>, <bos>, <pad>, :, S, P, C, A, I)\\
Hidden size ($d_{\text{model}}$) & 128 \\
Intermediate (FFN) size ($d_{\text{ff}}$) & 512 \\
Number of layers ($L$) & 4 \\
Number of attention heads ($h$) & 2 \\
Number of key/value heads & 2 \\
Head dimension ($d_k = d_{\text{model}}/h$) & 64 \\
Attention groups ($h / \text{KV heads}$) & 1 \\
Activation function & SwiGLU (SiLU Gating) \\
Positional embeddings & Rotary (RoPE), $\theta=10000$ \\
Max position embeddings & 32768 \\
Attention dropout & 0.0 \\
Normalization & RMSNorm, $\epsilon=10^{-6}$ \\
Initializer range & 0.02 \\
Sliding window & Disabled (Window Size = 4096) \\
Tie word embeddings & False \\
\bottomrule
\end{tabular}
\end{table}

\subsection{Data}
Here, we explain the details of the synthetic data generated in our study in Sec.~\ref{sec:syn_math}. The framework is designed to create a controllable environment that mimics critical aspects of complex problem-solving, such as deducing a sequence of operations and handling spurious correlations or irrelevant information. The core task involves navigating a discrete state space by applying actions, where the correct action sequence must be inferred from an auxiliary sequence of ``problem states.''

The synthetic environment is built upon several key entities, visually represented in Fig.~\ref{fig:syn_set_up}~{a)} of the main text:

\begin{itemize}
    \item \textbf{States (\(S_x\))}: These represent the fundamental state of the problem instance. The model's goal is to correctly predict the trajectory of these states. Each state \(s\) is an integer from the set \(\{0, 1, \dots, N_S-1\}\), where \(N_S\) is the total number of unique states.
    \item \textbf{Actions (\(A_x\))}: These are discrete operations that cause transitions between states \(S_x\). Each action \(a\) is an integer from the set \(\{0, 1, \dots, N_A-1\}\), where \(N_A\) is the total number of unique actions.
    \item \textbf{Problem States (\(P_x\))}: This sequence of tokens serves as an indirect instruction or ``problem description.'' The model must learn to interpret the transitions between problem states to infer the appropriate action \(A_x\) to apply to the current state \(S_x\). Each problem state \(p\) is an integer from the set \(\{0, 1, \dots, N_P-1\}\), where \(N_P\) is the total number of unique problem states.
    \item \textbf{Spurious Contexts (\(C_x\))}: These are tokens associated with each chosen action \(A_x\). Their statistical properties are deliberately manipulated between different data splits:
        \begin{itemize}
            \item During \textbf{pre-training} (the `train' split), the context \(c\) for a given action \(a\) is sampled from a non-uniform distribution \(P_{\beta}(c|a)\). This distribution is specifically engineered to create strong, yet spurious, correlations between certain contexts and actions, acting as a potential distractor or misleading heuristic for the model. We specify this distribution in the next paragraph.
            \item During \textbf{Reinforcement Learning (RL) training} (the `train\_rl' split) and \textbf{evaluation} (the `eval' split), these spurious correlations are removed. The context \(c\) for a given action \(a\) is sampled from a uniform distribution, \(P(c|a) = 1/N_C\), where \(N_C\) is the total number of unique context tokens.
        \end{itemize}
    \item \textbf{Irrelevant Contexts (\(I_x\))}: A fixed number of randomly generated tokens are appended to the input prompt. These tokens have no causal relationship with the problem's solution and are included to test the model's ability to ignore irrelevant information and focus on salient parts of the input. We specify the parametrization in the next paragraph.
\end{itemize}

The data generation process is controlled by a set of parameters. Their default values are as follows:
\begin{itemize}
    \item \(n_{\text{dim}}\) (Default: 20): This parameter defines the dimensionality of the primary entities. Specifically, the number of distinct actions \(N_A\) is set to \(n_{\text{dim}}\).
    \item \(n_{\text{ctx}}\) (Default: 20): This parameter sets the size of other entity spaces. For all experiments explored, we set the number of distinct states \(N_S\), the number of distinct problem states \(N_P\), and the number of distinct context tokens \(N_C\) all equal to \(n_{\text{ctx}}\).
        Thus, by default: \(N_A = 20\), \(N_S = 20\), \(N_P = 20\), \(N_C = 20\).
    \item \(n_{\text{irr\_ctx}}\) (Default: 10): Specifies the quantity of irrelevant context tokens (`num\_irr\_context') to be included in each problem prompt.
    \item \(k_{\text{min}}\) (Default: 5): The minimum number of actions (`min\_actions\_per\_problem`) that will constitute a single problem instance.
    \item \(k_{\text{max}}\) (Default: 10): The maximum number of actions (`max\_actions\_per\_problem') for a single problem instance. The actual number of actions for each problem is chosen uniformly at random from the range \([k_{\text{min}}, k_{\text{max}}]\).
    \item \(\alpha_{\text{train}}\) (Default: 0.3): This is the concentration parameter for the Dirichlet distribution used to generate the action probabilities \(P_{\alpha}(A_i|S_i)\) for the \textbf{pre-training (`train') split}. Specifically, for each state \(s\), the probability distribution over actions is drawn as \(\text{Dirichlet}(\vec{\alpha})\), where \(\vec{\alpha}\) is a vector of length \(N_A\) with all elements equal to \(\alpha_{\text{train}}\). A value less than 1 (like 0.3) tends to produce distributions where a few actions are highly probable from a given state, introducing bias.
    \item \(\beta_{\text{train}}\) (Default: 0.3): This is a concentration parameter for the Dirichlet distribution used to generate the spurious context probabilities \(P_{\beta}(C_i|A_i)\) for the \textbf{pre-training (`train') split}. Here, for each action \(a\), the probability distribution over contexts is drawn as \(\text{Dirichlet}(\vec{\beta})\), where \(\vec{\beta}\) is a vector of length \(N_C\) with all elements equal to \(\beta_{\text{train}}\). This creates non-uniform associations between actions and contexts.
\end{itemize}

For the \textbf{RL (`train\_rl') and eval (`eval') splits}, the action selection is uniform random (i.e., \(P(A_i|S_i) = 1/N_A\)), and context generation is also uniform random (i.e., \(P(C_i|A_i) = 1/N_C\)). This ensures that these splits are free from the biases intentionally introduced in the pre-training data.

The evolution of states and problem states is governed by deterministic transition tables, while the selection of actions and contexts in the pre-training phase is probabilistic:

\begin{enumerate}
    \item \textbf{State Transition Table (\(\mathcal{T}_S\))}: This is a matrix of dimensions \(N_S \times N_A\). An entry \(\mathcal{T}_S(s, a)\) specifies the next state \(s'\) resulting from taking action \(a\) in state \(s\). This table, `state\_transitions`, is initialized by drawing each entry uniformly at random from \(\{0, \dots, N_S-1\}\).

    \item \textbf{Problem State Transition Table (\(\mathcal{T}_P\))}: This is a matrix of dimensions \(N_P \times N_A\). An entry \(\mathcal{T}_P(p, a)\) specifies the next problem state \(p'\) when action \(a\) is taken while the current problem state is \(p\). The table `problem\_state\_transitions' is constructed such that for any given current problem state \(p\), applying different actions leads to unique next problem states. This is achieved by ensuring that for each row \(p\), the entries \(\mathcal{T}_P(p, \cdot)\) consist of \(N_A\) distinct values, drawn from a permutation of \(\{0, \dots, N_P-1\}\).

    \item \textbf{Action Probability Table for Pre-training (\(P_{\text{action}}\))}: This table, `action\_probs', has dimensions \(N_S \times N_A\). Each row \(s\) contains the probability distribution \(P_{\alpha}(A|S=s)\) over actions, generated from \(\text{Dirichlet}(\alpha_{\text{train}} \cdot \vec{1}_{N_A})\). This table is only used for action selection in the `train' split. For the `train\_rl' and `eval' splits, actions are chosen uniform randomly.

    \item \textbf{Action-Context Probability Table for Pre-training (\(P_{\text{context}}\))}: This table, `action\_context\_prob\_train', has dimensions \(N_A \times N_C\). Each row \(a\) contains the probability distribution \(P_{\beta}(C|A=a)\) over contexts, generated from \(\text{Dirichlet}(\beta_{\text{train}} \cdot \vec{1}_{N_C})\). This table is only used for context generation in the `train' split.

    \item \textbf{Action-Context Probability Table for RL/Eval}: For the `train\_rl' and `eval' splits, the context \(c_t\) for an action \(a_t\) is chosen uniformly. This is represented by `action\_context\_prob\_test', an \(N_A \times N_C\) matrix where each entry is \(1/N_C\).
\end{enumerate}

Each individual data instance, comprising a prompt (the ``question") and a completion (the ``solution"), is constructed through the following sequence of operations:

\begin{enumerate}
    \item \textbf{Determine Trajectory Length}: The number of actions, \(k\), for the current instance is selected by drawing an integer uniformly from the interval \([k_{\text{min}}, k_{\text{max}}]\). This \(k\) defines the length of the action sequence and consequently the number of state transitions.

    \item \textbf{Initialize Sequences}:
        \begin{itemize}
            \item An initial state \(s_0\) is chosen uniformly at random from \(\{0, \dots, N_S-1\}\).
            \item An initial problem state \(p_0\) is chosen uniformly at random from \(\{0, \dots, N_P-1\}\).
            \item Empty lists are prepared for actions (\(\mathcal{A}\)), states (\(\mathcal{S} = [s_0]\)), problem states (\(\mathcal{P} = [p_0]\)), and contexts (\(\mathcal{C}\)).
        \end{itemize}

    \item \textbf{Generate Trajectory Core (Iterate \(k\) times, for \(t = 0, \dots, k-1\))}:
        \begin{itemize}
            \item \textbf{Action Selection (\(a_t\))}:
                \begin{itemize}
                    \item \textbf{For `train' split}: The action \(a_t\) is sampled based on the current state \(s_t\) using the pre-calculated biased probabilities: \(a_t \sim P_{\text{action}}(A | S=s_t)\).
                    \item \textbf{For `train\_rl' and `eval' splits}: The action \(a_t\) is sampled uniformly at random from the set of all possible actions \(\{0, \dots, N_A-1\}\).
                \end{itemize}
                The chosen action \(a_t\) is appended to \(\mathcal{A}\).

            \item \textbf{Determine Next State (\(s_{t+1}\))}: The next state \(s_{t+1}\) is obtained deterministically using the state transition table: \(s_{t+1} = \mathcal{T}_S(s_t, a_t)\). This \(s_{t+1}\) is appended to \(\mathcal{S}\).

            \item \textbf{Determine Next Problem State (\(p_{t+1}\))}: The next problem state \(p_{t+1}\) is obtained deterministically using the problem state transition table: \(p_{t+1} = \mathcal{T}_P(p_t, a_t)\). This \(p_{t+1}\) is appended to \(\mathcal{P}\).

            \item \textbf{Context Generation (\(c_t\))}:
                \begin{itemize}
                    \item \textbf{For `train' split}: The context token \(c_t\) is sampled based on the chosen action \(a_t\) using the pre-calculated spurious correlation probabilities: \(c_t \sim P_{\text{context}}(C | A=a_t)\).
                    \item \textbf{For `train\_rl' and `eval' splits}: The context token \(c_t\) is sampled uniformly at random from the set of all possible context tokens \(\{0, \dots, N_C-1\}\).
                \end{itemize}
                The chosen context \(c_t\) is appended to \(\mathcal{C}\).
        \end{itemize}
    After \(k\) iterations, we have: actions \((a_0, \dots, a_{k-1})\), states \((s_0, \dots, s_k)\), problem states \((p_0, \dots, p_k)\), and contexts \((c_0, \dots, c_{k-1})\).

    \item \textbf{Generate Irrelevant Context Tokens}: A sequence of \(n_{\text{irr\_ctx}}\) irrelevant tokens, \((i_0, \dots, i_{n_{\text{irr\_ctx}}-1})\), is generated by sampling each token uniformly at random from \(\{0, \dots, n_{\text{dim}}-1\}\). Note: `n\_dim' is used here as per the code `np.random.randint(0,n\_dim,size=num\_irr\_context)'.

    \item \textbf{Construct Prompt and Completion Strings}:
        \begin{itemize}
            \item \textbf{Prompt String (Input to Model)}:
            The prompt is assembled by concatenating string representations of the generated tokens.
            It starts with the initial state: \texttt{"S <\(s_0\)>"}
            Then, for each step \(j\) from \(0\) to \(k\):
                Append problem state: \texttt{"P <\(p_j\)>"}
                If \(j < k\) (i.e., not the last problem state), append context: \texttt{"C <\(c_j\)>"}
            Append irrelevant context tokens: \texttt{"I <\(i_0\)> I <\(i_1\)> \dots I <\(i_{n_{\text{irr\_ctx}}-1}\)>"}
            End with a separator: \texttt{":"}

            The structure is:
            \texttt{S <\(s_0\)> P <\(p_0\)> C <\(c_0\)> P <\(p_1\)> C <\(c_1\)> \dots P <\(p_{k-1}\)> C <\(c_{k-1}\)> P <\(p_k\)> I <\(i_0\)> \dots I <\(i_{n_{\text{irr\_ctx}}-1}\)> : }

            \item \textbf{Completion String (Target Output for Model)}:
            The completion represents the desired step-by-step solution.
            For each step \(j\) from \(0\) to \(k\):
                Append state: \texttt{"S <\(s_j\)>"}
                If \(j < k\), append action: \texttt{"A <\(a_j\)>"}
            End with a padding token: \texttt{"<pad>"}

            The structure is:
            \texttt{S <\(s_0\)> A <\(a_0\)> S <\(s_1\)> A <\(a_1\)> \dots S <\(s_{k-1}\)> A <\(a_{k-1}\)> S <\(s_k\)> <pad>}
        \end{itemize}
\end{enumerate}

This generation pipeline produces datasets (`train', `train\_rl', `eval') with controlled statistical properties. The `train' set has biased action priors and spurious action-context correlations, designed to reflect complexities and pitfalls in real-world pre-training data. The `train\_rl' and `eval' sets, by contrast, has uniform action and context sampling more suitable for thorough evaluation of evaluating generalization beyond learned spurious cues. This latter also provides a cleaner signal for RL.

We use a default of $200,000$ pre-training sequences, $128$ RL sequences and $512$ eval sequences.

\subsection{Training}
We use the \texttt{Transformers} \citep{wolf2020huggingfacestransformersstateoftheartnatural} library to pre-train the transformer described in App.~\ref{appdx:synthetic_setup_model} with autoregressive next token cross entropy loss. The training hyperparameters are described in Table~\ref{tab:sft_hyperparams}. For RL, we use \texttt{verl} \citep{Sheng2024-rs}'s GRPO pipeline on the pre-trained model. The RL hyperparameters are in Table~\ref{tab:synth_rl_hyperparams}.

\begin{table}[h]
    \centering
    \caption{Hyperparameters used in pre-training.}\label{tab:sft_hyperparams}
    \vspace{10pt}
    \begin{tabular}{@{}llr@{}}
        \toprule
        \textbf{Category} & \textbf{Hyperparameter} & \textbf{Value} \\
        \midrule
        \textbf{Data Configuration}
        & use\_assistant\_mask      & \texttt{false} \\
        & train\_max\_length        & 128 \\
        \midrule
        \textbf{Training Configuration} 
        & batch\_size       & 64 \\
        & num\_train\_epochs         & 1 \\
        & bf16                      & \texttt{false} \\
        \midrule
        \textbf{Optimizer Configuration} 
        & learning\_rate            & 0.001 \\
        & lr\_scheduler\_type       & \texttt{linear} \\
        & warmup\_ratio              & 0.05 \\
        & gradient\_accumulation\_steps & 1 \\
        \bottomrule
    \end{tabular}
\end{table}

\begin{table}[h]
    \centering
            \caption{Hyperparameters used in RL}\label{tab:synth_rl_hyperparams}
        \vspace{10pt}
        \begin{tabular}{@{}llr@{}}
        \toprule
        \textbf{Category} & \textbf{Hyperparameter} & \textbf{Value} \\
        \midrule
        \textbf{Data} 
        & max\_prompt\_length      & 64 \\
        & max\_response\_length      & 64 \\
        & train\_batch\_size         & 32 \\
        \midrule
        \textbf{Actor} 
        & ppo\_mini\_batch\_size     & 32 \\
        & grad\_clip                & 0.2 \\
        & clip\_ratio               & 0.2 \\
        & entropy\_coeff            & 0.001 \\
        & use\_kl\_loss              & True \\
        & kl\_loss\_coef             & 0.001 \\
        & kl\_loss\_type             & low\_var\_kl \\
        & ppo\_epochs               & 4 \\
        \midrule
        \textbf{Optimizer} 
        & lr                         & 5e-5 \\
        & lr\_warmup\_steps\_ratio   & 0.0 \\
        & warmup\_style              & constant \\
        \midrule
        \textbf{Rollout} 
        & temperature               & 1.0 \\
        & top\_k                     & Not Used \\
        & top\_p                     & 1 \\
        & response\_length          & 64 \\
        & n\_rollout\_samples                         & 8 \\
        \midrule
        \textbf{Algorithm}
        & kl\_coef                  & 0.001 \\
        & adv\_estimator             & GRPO \\
        \midrule
        \textbf{Trainer} 
        & total\_epochs             & 100 \\
        & save\_freq                & 10 \\
        & validation\_freq                & 5 \\
        \bottomrule
        \end{tabular}
\end{table}

\paragraph{Computations Resouces} We used NVIDIA A100 and NVIDIA H100 GPUSs for the synthetic experiment. Pretraining for $\sim3000$ steps took roughly 3 minutes on a single GPU and RL for $~\sim300$ steps took roughly 10 minutes on a single GPU.


\end{document}